\documentclass[sigconf]{acmart}
\usepackage{algorithm}
\usepackage{algorithmic}
\usepackage{amsmath}
\usepackage{graphicx}
\usepackage{array}
\usepackage{pdflscape}
\usepackage{booktabs}
\usepackage{tabularx}
\usepackage{amsfonts}

\DeclareMathOperator*{\argmax}{arg\,max}

\AtBeginDocument{%
  \providecommand\BibTeX{{%
    \normalfont B\kern-0.5em{\scshape i\kern-0.25em b}\kern-0.8em\TeX}}}

\copyrightyear{2020}
\acmYear{2020}
\setcopyright{acmcopyright}\acmConference[ICAIF '20]{ACM International Conference
on AI in Finance}{October 15--16, 2020}{New York, NY, USA}
\acmBooktitle{ACM International Conference on AI in Finance (ICAIF '20), October
15--16, 2020, New York, NY, USA}
\acmPrice{15.00}
\acmDOI{10.1145/3383455.3422547}
\acmISBN{978-1-4503-7584-9/20/10}

\begin{document}

\title{SURF: Improving classifiers in production by learning from busy and noisy end users}


\author{Joshua Lockhart}
\email{joshua.lockhart@jpmorgan.com}
\affiliation{%
  \institution{J.P. Morgan AI Research}
}
\author{Samuel Assefa}
\affiliation{%
  \institution{J.P. Morgan AI Research}
}
\author{Ayham Alajdad}
\affiliation{%
  \institution{J.P. Morgan Applied AI \& ML}
}
\author{Andrew Alexander}
\affiliation{%
  \institution{J.P. Morgan Applied AI \& ML}
}

\author{Tucker Balch}
\affiliation{%
  \institution{J.P. Morgan AI Research}
}
\author{Manuela Veloso}
\affiliation{%
  \institution{J.P. Morgan AI Research}
}

\renewcommand{\shortauthors}{Lockhart et al.}

\begin{abstract}
Supervised learning classifiers inevitably make mistakes in production, perhaps mis-labeling an email, or flagging an otherwise routine transaction as fraudulent. It is vital that the end users of such a system are provided with a means of relabeling data points that they deem to have been mislabeled. The classifier can then be retrained on the relabeled data points in the hope of performance improvement. To reduce noise in this feedback data, well known algorithms from the crowdsourcing literature can be employed. However, the feedback setting provides a new challenge: how do we know what to do in the case of user non-response? If a user provides us with no feedback on a label then it can be dangerous to assume they implicitly agree: a user can be busy, lazy, or no longer a user of the system! We show that conventional crowdsourcing algorithms struggle in this user feedback setting, and present a new algorithm, SURF, that can cope with this non-response ambiguity.
\end{abstract}
\maketitle
\section{INTRODUCTION}
In this era of enterprise data science, supervised learning systems have become ubiquitous in large organisations: categorising emails, detecting fraudulent transactions, deciding which job applicants to invite to interview, \emph{etc}.
When building and deploying such systems, the classification accuracy when the model is run in a sanitised testing environment is only part of the story. When the algorithm is deployed, substantial effort needs to go into ensuring it performs well in a non-stationary environment \emph{e.g.} the distributions of the classes can change in time, the user requirements may change, or perhaps even the original training data provided by the users is missing some part of the problem taxonomy.

A useful tool in the arsenal of an enterprise data scientist is the \emph{feedback mechanism}: provide end users with a mechanism to \emph{relabel} any data point they receive. The relabeled data is sent to the team maintaining the classifier, who use collections of these relabeled data points to retrain the model with the aim of improving future performance.

Multiple interesting problems arise in this `feedback loop' setting. The first is that of noisy labels. Not all users should be trusted, since some may not understand the meaning of some subset of the label taxonomy, while others may be resistant to automation and provide intentionally incorrect relabelings to the system as a form of protest. A classifier with a feedback loop must be able to identify which users can be trusted to provide correct and useful relabelings. 
The problem becomes more subtle when we realise that what may look like noise to a machine learning classifier may in fact be signal: a classifier must learn to be corrected by its users, and not just to correct them when it deems them to be wrong. 

Identifying and filtering out naive, and adversarial, noise in user provided labeled data has enjoyed a great deal of attention lately with the rise of crowdsourcing platforms such as Amazon Mechanical Turk. However, the crowdsourcing of \textit{feedback} labels as described above, where we are provided with user relabelings of a data point that are provided \textit{in response to} a classifier's label for that data point, is a different problem to the conventional crowdsourcing setting where labels are provided \textit{ex nihilo} by crowd workers.

Indeed, considering the feedback loop problem as a simple crowdsourcing problem overlooks a key element of how feedback is collected: if a user receives a data point assigned a label that they agree with, then they will not relabel it. From this aspect, ambiguity arises. Consider an end user who is particularly busy with their normal workload for the day, who receives an email that has been incorrectly labeled by the machine learning system. In this case, the user may never give a relabeling of the incorrectly labeled email, which will simply sit in their inbox. The assumption that \textit{no response} from an end user should be taken as tacit agreement with the label provided by the classifier will lead to an erroneous judgement being made. Potentially, the classifier could be retrained with incorrect labeled data and its performance could get worse.

In this work we present an algorithm that takes data points that have been labeled by a classifier, along with relabelings provided by users, and produces an estimate of the ground truth of those data points. The algorithm, SURF (\textbf{S}elective \textbf{U}se of use\textbf{R} \textbf{F}eedback), is an extension of the well known Dawid-Skene algorithm \cite{DawidSkene}, which is based on the principle of expectation maximisation (EM). 
Specifically, Dawid-Skene estimates a \emph{confusion matrix} for each crowd worker (in our case, the end users, along with the actual production classifier), and uses this to improve its estimate of the actual ground truth labels. We extend this algorithm by providing a means of estimating the response rate of each end user, meaning we can distinguish between diligent end users whose non-response should be interpreted as agreement, and busy / lazy end users whose non-response should be regarded as `no label'.

We show that by augmenting Dawid-Skene's EM loop with this estimate of what we (charitably) refer to as user \emph{busyness}, the algorithm converges to a better estimate of the ground truth when users are noisy. We empirically demonstrate this by building a multi-agent simulation of a production environment, in which simulated end user agents receive labeled data points from the MNIST data set, and provide noisy relabelings when they deem the classifier to be incorrect. The end user agents each have a \textit{busyness probability}: if they are busy then they will not respond with feedback, and their label is deemed to be equal to the label they received from the classifier.

We demonstrate that our algorithm (SURF) is able to cope with user non-response more effectively than a number of algorithms drawn from the crowdsourcing literature, suggesting that taking this into account is vital in production systems.

\subsection{RELATED WORK}
A number of probabilistic algorithms have been brought to bear on the problem of inferring true labels from noisy crowdsourced labels. In an early work, Dawid and Skene \cite{DawidSkene} treat each crowd worker as a noisy channel through which the ground truth labels pass. Thus, they learn a confusion matrix for each worker, and use this to estimate ground truth. The Bayesian classifier combination (BCC) algorithms of Kim and Ghahrami \cite{BCC} take this confusion matrix view and apply it to the problem of combining labels given by \textit{ensembles of classifiers}. Explicitly, they provide algorithms for the case where classifiers provide labels independently from one another (IBCC), as well as the case where some classifiers behave in a way that their resultant labels are correlated (DBCC).
These ideas are utilised and extended by Simpson et al. \cite{simpson2013dynamic} for crowdsourcing problems proper. Simpson et al. cluster workers based on their behaviours, an idea also considered by Venanzi et al. \cite{community}.
Li et al. \cite{li2019exploiting} also consider correlated workers, again building on the BCC framework. They argue that DBCC \cite{BCC} does not scale well to the crowdsourcing setting, which may have a large number of crowd workers. They forgo the Markov network used in DBCC to estimate correlations between classifiers and use tensor decomposition instead, proposing \emph{enhanced Bayesian Classifier combination} (EBCC).

Alternative techniques to estimating confusion matrices include modeling worker \textit{ability} and datapoint \textit{difficulty} as in Whitehall et al.'s GLAD (Generative model of Labels, Abilities, and Difficulties), \cite{Whitehill2009WhoseVS}, the minimax entropy approach of Zhou et al. \cite{zhou2012learning}, and asking workers to provide labeling functions as in the Data Programming paradigm of Ratner et al.
\cite{Ratner2016DataPC}. Welinder et al. \cite{Welinder2010TheMW} provide another Bayesian treatment of the problem, again considering worker competence rather than a confusion matrix approach. 

We recognise that none of the algorithms described above explicitly tackle the feedback setting we consider in this paper, which to the best of our knowledge has not been considered from a crowdsourcing point of view before. Thus, the fact that our algorithm, SURF, does much better than these algorithms at this `busy feedback' task is perhaps not surprising. Nevertheless, our experiments encompass what we feel would be the natural tools that data scientists would turn to in order to do crowdsourcing, and as we will see, our experiments show that they are not performant for the task we are interested in.
\section{METHODS}
\subsection{SIMULATING USERS THAT GIVE FEEDBACK TO A SYSTEM}
We explicitly simulate the production setting described earlier in order to perform our experiments. We begin with a set of data points drawn at random from the MNIST data set \cite{MNIST}. The labels of these data points are referred to as the \textit{ground truth}. For each data point in this set we obtain a new label from a \textit{classifier agent}. We send these labeled data points (the \textit{classifier labels}) to a fixed pool of $K$ \emph{user agents}, who will provide \emph{feedback} on the classifier labels. The issue we face is that the user agents can be noisy, providing erroneous relabelings. The task is to identify which labels are erroneous, ideally achieving a good estimate of the ground truth labels. 

The interpretation here is that the classifier in production attempts to label the MNIST data points, potentially making mistakes. The data science team hope to augment their training data set with the feedback from the end users, but first they must remove noise injected by the classifier itself, and the user agents.


We consider user agents that make errors at the label level rather than the data point level: each user is parametrized by a probabilistic function on the label set $\mathcal{L}=\{1,\dots,L\}$ of the classification problem. That is, a pool of user agents is parametrized by a set of probabilistic functions $\{u_i:\mathcal{L}\rightarrow\mathcal{L}\}_i$. Feedback on a data point labeled by the classifier $y_i^{(0)}$ is then obtained by passing the ground truth of that data point through the user pool error functions $\{y_i^{(1)},\dots, y_i^{(K)}\} := \{u_1(y_i),\dots, u_K(y_i)\}$.

The user agents we have described so far are sufficient to simulate the \emph{ex nihilo} crowdsourcing setting where we take a list of user provided labels and attempt to distil some ground truth. The user agents don't need to take the classifier label into account at all, they are simply a noisy channel through which the ground truth passes.

This is however not sufficient for us to model the feedback setting we are interested in, where the label provided by users for a particular data point is prompted by a classifier's label for that data point. The particular aspect of this setting we are concerned with is the inherent ambiguity of a user's non-response to a label. Specifically, it is not clear if we should interpret a non-response as an implicit agreement with that label (in which case the label provided by the user matches the label it was requested to give feedback on). If a user is too busy to give feedback on an incorrect label then such an assumption of agreement-by-silence is too strong, we do not know what label the user would have provided if they were not busy and fully engaged with the feedback process.

For this purpose, we define an $\epsilon$-busy user agent as a function $u_\epsilon: \mathcal{L}\rightarrow\mathcal{L}$, such that for any label from the classifier $y_i^{(0)}$ we have that
\begin{align*}
u_\epsilon(y_i^{(0)}) = \begin{cases}
y_i^{(0)}&\text{ with probability }\epsilon\\
u(y_i)&\text{ with probability } 1-\epsilon,\\
\end{cases}
\end{align*}
where $u:\mathcal{L}\rightarrow\mathcal{L}$ is a standard user agent error function as defined above. Literally, an $\epsilon$-busy user will respond with its opinion on the data point's true label with probability $1-\epsilon$, and will respond with the classifier label otherwise (for it is busy).

The task we consider in this paper is how to take a classifier's labels for a set of data points, along with the user pool's feedback on those labels, and produce a new set of labeled data points that can be used to train a new classifier with better accuracy. Explicitly, we are interested in taking the labels $\{y_i^{(0)},\dots, y_i^{(K)}\}_i$, and mapping them to some $\hat{y}_1,\dots, \hat{y}_N$ that are a good approximation of the ground truth labels $y_1,\dots, y_N$.

The labels provided as feedback from busy users are correlated with the classifier labels. As we will see, tools that assume users are independent \emph{e.g.} the Dawid-Skene algorithm, perform poorly in the feedback setting. In the next section we formulate algorithms that take this correlation into account by estimating each user's busyness parameter. If we have a handle on this parameter then we can build a better estimate of the noise that each user provides feedback with, and thus obtain a better approximation of the ground truth of the problem.

\subsection{ALGORITHMS}
As discussed in the previous section, the algorithms we consider and build upon in this paper attempt to estimate a list of ground truth labels $y_1,\dots,y_N$, given labels $\{y_i^{(1)},\dots, y_{i}^{(K)}\}_i$ provided by a fixed set of $K$ users, in response to a classifier's labels $y_1^{(0)}\dots y_N^{(0)}$ of those same data points. In what follows, it will make the notation easier to follow if we treat the classifier as a user. Specifically, we refer to the classifier as agent $u_0$, then define the set $U=\{u_0,\dots, u_K\}$ to encompass all users along with the classifier. We occasionally refer to the labels $y_i^{(0)},\dots, y_i^{(K)}$ as $Y_i^U$ for short.

The simplest way of estimating each label $y_i$ would be to take the majority vote over the user provided labels for that data point \begin{align*}\text{maj}(y_i^{(0)},\dots,y_i^{(K)}).\end{align*} However, if the users are noisy in predictable ways then it is possible to do better. The Dawid-Skene (DS) algorithm \cite{DawidSkene} instead estimates a probability distribution over labels for each data point
\begin{align}
    p(y_i=l~|~Y_i^U).\label{eq:pdist}
\end{align}
Once a reasonable estimate has been obtained of this data point, a data scientist could train a supervised model with the data points labeled by either \begin{align*}\argmax_l p(y_i=l~|~Y_i^U),\end{align*}
or with the probability vector \begin{align*}
    \textbf{y}_i^U = \begin{pmatrix}p(y_i=0~|~Y_i^U), \dots, p(y_i=L~|~Y_i^U) \end{pmatrix}.\end{align*}
Dawid and Skene estimate the probability in Eq. \ref{eq:pdist} by making the assumption that each user makes errors at the level of the labels, rather than at the level of the data points themselves. This is the same as in our user agent model described above: a user agent makes errors specified by a probabilistic function acting on labels $u:\mathcal{L}\rightarrow\mathcal{L}$, and when queried about the label of the data point with index $i$ will respond with the true label of that data point, acted on by that probabilistic function $a(i)=u(y_i)$. This way, with each user $u_k\in U$ we can associate a matrix of probabilities \begin{align*}\pi_{lm}^{(k)}=p(y^{(k)}=m~|~y=l),\end{align*} to be literally interpreted as the probability a user confuses data points labeled $l$ with data points labeled $m$. We refer to these as the user's \textit{confusion probabilities}.

Assuming users report their labels independently of one another, and independently of the data point index, Dawid and Skene obtain

\begin{align*}
    p(y_i=l~|~y_i^{(1)}&=l_1,\dots, y_i^{(K)}) \\
    &\propto p(y_i=l)\prod_k p(y_i^{(k)}=l_k~|~y_i=l)\\
    &= q_l\prod_k \pi_{l,l_k}^{(k)}.
\end{align*}
The quantities in this expression can be estimated via expectation maximisation (EM). To do so, Dawid and Skene define the following
\begin{align}
\hat{\pi}_{lm}^{(k)} = \frac{\sum_j \mathbb{I}(y_j^{(k)}=l)\mathbb{I}(\hat{y}_j = m)}{\sum_{l'} \sum_j \mathbb{I}(y_j^{(k)}=l')\mathbb{I}(\hat{y}_j = m)},\label{eq:dscmat}
\end{align}
\begin{align}
\hat{q}_l=\frac{\sum_j\mathbb{I}(\hat{y}_j = l)}{\sum_{l'}\sum_j\mathbb{I}(\hat{y}_j = l')},\label{eq:dsprob}
\end{align}
where $\mathbb{I}(\cdot)$ is an indicator variable equal to $1$ if the condition given as its argument is true and $0$ otherwise \footnote{Note that using indicator variables here is a simplified version of the algorithm. If desired, these can be reweighted by that iteration's estimates of label probabilities \emph{etc.}}. 

Finally, 
\begin{align}
\hat{p}(y_i=l~|~y_i^{(0)}=l_{0},\dots, y_i^{(K)}=l_{K})\nonumber\\ =\frac{\hat{q}_l\prod_{k=0}^{K} \hat{\pi}_{l_j,l}^{(k)}}{\sum_{l'}\hat{q}_{l'}\prod_{k=0}^{K} \hat{\pi}_{l_j,l'}^{(k)}}\label{eq:dspost}
\end{align}
These estimators are utilised by Dawid and Skene in Algorithm \ref{alg:ds}\footnote{Two different implementations of this algorithm are at \url{https://github.com/dallascard/dawid_skene}, and \url{https://github.com/zhydhkcws/crowd_truth_infer}.}.
\begin{algorithm}[ht!]
    \caption{\textsc{Dawid-Skene} \cite{DawidSkene}}
    \label{alg:ds}
    \begin{algorithmic}
        \FOR{$i=1$ {\bfseries to} $M$}
            \STATE initialize $\hat{y}_i$
        \ENDFOR
        \WHILE{termination condition not met}
        \STATE update $\hat{\pi}^{(0)}, \dots, \hat{\pi}^{(K)}$ via Eq. \ref{eq:dscmat}
        \STATE update entries of $\hat{\textbf{q}}$ via Eq. \ref{eq:dsprob}
        \FOR{$i=1$ {\bfseries to} $M$, $j=1$ {\bfseries to} $L$}
            \STATE update $\hat{p}(y_i=j|~y_i^{(0)},\dots, y_i^{(K)})$ via Eq. \ref{eq:dspost}
        \ENDFOR
        \ENDWHILE
    \end{algorithmic}
\end{algorithm}

Algorithm \ref{alg:ds} can be utilised for our purposes if we treat the classifier labels and the user feedback on those labels as $Y^{U}$. However, as we show later, such an algorithm will run into problems when we have a user pool with busy users.
Users who are busy provide labels that are correlated with the classifier labels. The independence assumption on the users means that this algorithm is not sufficient to model how users give feedback.
Our contribution is a new EM based algorithm that estimates each user's busyness, and utilises this in its estimate of $p(y|U)$.

Explicitly, suppose we knew the busyness parameter for a particular user in the pool, $\epsilon_k$. We know that with probability $\epsilon_k$, such a user will simply respond with the classifier's label. We can make a better estimate of that user's confusion probabilities by taking this into account in the following way
\begin{align}
\hat{\pi}_{l,m,m'}^{(k)} =\begin{cases}
                                             \hat{\epsilon}_k + (1-\hat{\epsilon}_k)\hat{\pi}_{l,m}^{(k)} & \text{ if } m = m'\\
                                             \hat{\pi}_{l,m}^{(k)} & \text{ otherwise. }
                                             \end{cases}\label{eq:surfcmat}
\end{align}
That is, if a user were to respond with the same label as the classifier gave for a data point, then there is the possibility that they are busy and have not responded. On the other hand, if that user gives a totally separate label to what the classifier gave, we treat this as a label like any other.

Estimating the posterior distribution over labels is thus a function of these conditional confusion probabilities 
\begin{align}
\hat{p}(y_i=l~|~y_i^{(0)}=l_{0},\dots, y_i^{(K)}=l_{K})\nonumber\\ = \frac{\hat{q}_l \hat{\pi}_{l,l_0}^{(0)}\prod_{k=1}^{K} \hat{\pi}_{l,l_k,l_0}^{(k)}}{\sum_{l'}\hat{q}_l \hat{\pi}_{l',l_0}^{(0)}\prod_{k=1}^{K} \hat{\pi}_{l',l_k,l_0}^{(k)}}.\label{eq:surfpost}
\end{align}
Estimating each user's busyness probability can be done by taking into account the estimates of the posterior probabilities. Intuitively, suppose a user tends to give labels that agree with the classifier, even when we estimate the classifier labels to be of low accuracy. Such a user is likely to be a highly busy user that rarely engages with the feedback process. Explicitly, suppose our estimate of the posterior tells us that the classifier's label is likely to be wrong, that is, $\hat{p}(y_i = y_i^{(0)}|U)$ is low. If a user provides a label for that same data point that agrees with the classifier's label, then it is likely the user is not engaging with the feedback process. 

This intuition can be used to obtain a maximum likelihood estimate of the parameters $\epsilon_1,\dots, \epsilon_K$. Whether a user's label $y_i^{(k)}$ agrees or disagrees with a classifier's label $y_i^{(0)}$ for a particular data point can be treated as a Bernoulli trial, where we refer to the success probability as the user acceptance probability $p_i^{(k)}$ for that data point. Whether the user was busy or not when they were queried about a data point is another Bernoulli trial, where the success probability is the user's busyness rate $\epsilon_k$. Then we have the likelihood
\begin{align*}
    L(\epsilon_k~|~y_i^{(0)},y_i^{(k)}) = \begin{cases} \epsilon_k + (1-\epsilon_k)p_{i}^{(k)}& \text{ if } y_{i}^{(0)}=y_{i}^{(k)}\\
    (1-\epsilon_k)(1-p_{i}^{(k)})& \text{ if } y_{i}^{(0)}\neq y_{i}^{(k)},
    \end{cases}
\end{align*}
and if we assume the user gives feedback on all data points independently
\begin{align}
\hat{\epsilon}_k = \argmax_{\epsilon} \prod_i L(\epsilon|y_i^{(0)},y_i^{(k)}).\label{eq:surfbusy}
\end{align}
In order to make this estimate, we assume that the probability a user will agree with the classifier label is exactly the probability the classifier's label is correct according to our estimate of the posterior distribution. Explicitly, we take $p_i^{(k)} = \hat{p}(y_i=y_i^{(0)}~|~ U)$. We incorporate these estimates into the EM loop to obtain our algorithm SURF. This is detailed in Algorithm \ref{alg:surf}, where we the variable $a_{ki}$ is $1$ if user $k$ was assigned data point $i$ and $0$ otherwise.

\begin{algorithm}[ht!]
    \caption{\textsc{SURF}}
    \label{alg:surf}
    \begin{algorithmic}
        \FOR{$i=1$ {\bfseries to} $N$}
        \STATE initialize $\hat{y}_i$
        \ENDFOR
        \FOR{$k=1$ {\bfseries to} $K$}
        \STATE initialize $\hat{\epsilon}_k = \frac{1}{N/M}\sum_{i=1}^{N} \mathbb{I}(y_i^{(k)} = y_i^{(0)})a_{ki}.$
        \ENDFOR
        \WHILE{termination condition not met}
        \STATE update $\hat{\pi}^{(0)}, \dots, \hat{\pi}^{(K)}$ via Eq. \ref{eq:surfcmat}
        \STATE update entries of $\hat{\textbf{q}}$ via Eq. \ref{eq:dsprob}
        \STATE update $\hat{p}(y_i=l~|~y_i^{(0)},\dots, y_i^{(K)})$ via Eq. \ref{eq:surfpost}
        \STATE update $\hat{\epsilon}_k$ via Eq. \ref{eq:surfbusy} and Eq. \ref{eq:surfpost}.
        \ENDWHILE
    \end{algorithmic}
\end{algorithm}
In the next section we investigate the performance of the algorithms above.

\section{EXPERIMENTS}
In order to investigate the performance of our algorithm, and other crowdsourcing techniques, we will carry out a simulation of a supervised learning classifier receiving feedback. We first fix a set of data points $x_1,\dots,x_N$ along with their ground truth labels $y_1,\dots, y_N$. In our case, we draw these from the well known MNIST dataset of handwritten digits \cite{MNIST}. From this set of data points we then obtain classifier labels $y_1^{(0)},\dots,y_N^{(0)}$, which are shared with a pool of user agents. 
The user agents provide us with feedback on the classifier labels. The task is to recover the ground truth labels $y_1,\dots,y_N$ from the classifier labels and the feedback labels provided by the user agents.

To model noise in the user agents, we make use of the \textit{pairwise-flipper} family of confusion probability matrices. Such matrices are parametrized by a error probability $p$, and for our purposes lead to users that with probability $1-p$ answer correctly, and with probability $p$ respond with an incorrect label. At the beginning of each experiment, each user is assigned a single such matrix to act as its confusion matrix. 

We are interested in examining the case where users give noisy feedback to a classifier, but also the converse scenario where users give clean feedback on a noisy classifier. In order to obtain control over the classifier performance used in our experiments, we model a classifier as an additional user agent. In this way, we can experiment directly with noisy classifiers.

Explicitly, the experimental procedure we execute is as follows

\begin{enumerate}
    \item Randomly select $N$ data points from MNIST, refer to these as $(x_1,y_1),\dots,(x_N,y_N)$.
    \item Initialise $K$ busy user agents $u_{\epsilon_1},\dots,u_{\epsilon_K}$ with confusion matrices drawn uniformly at random from the set of pairwise-flipper matrices with fixed error probability $p_u$.
    \item Initialise an additional agent with busyness probability $\epsilon=0$ and confusion matrix drawn uniformly at random from the pairwise-flipper matrices with fixed error probability $p_{c}$. This is the agent that we use to represent the classifier in production, we refer to this as the \emph{classifier agent} $u_c$.
    \item Obtain the classifier labels $y_1^{(0)},\dots,y_N^{(0)}=u_c(y_1),\dots,u_c(y_N)$.
    \item For each classifier label $y_i^{(0)}$, select $M$ user agents $u_{\epsilon_{k_1}},\dots,u_{\epsilon_{k_M}}$ uniformly at random from the user pool. From this subset of the pool, obtain feedback labels $y_i^{(k_1)},\dots,y_i^{(k_M)}$.
    \item Obtain estimates $\hat{y}_1,\dots,\hat{y}_N$ from the classifier labels and the feedback labels. Record accuracy $\text{Acc} = (1/N)\sum_i^{N}I(\hat{y}_i=y_i)$.
\end{enumerate}
In order to examine the performance of the crowdsourcing algorithms in full depth, we will vary a number of parameters specified above. Explicitly, we will vary $M$ the number of users that give feedback on each labeled datapoint, as well as the probability that the classifier makes a mistake $p_c$, as well as the probability that each user makes a mistake $p_u$ and the probability that users are busy $\epsilon$. We will consider only homogeneous user pools, that is, we keep $p_u$ and $\epsilon$ constant across users.

We will run SURF on labels obtained in the way described above. We will compare SURF to two groups of crowdsourcing algorithms: GROUP 1: Dawid-Skene (DS) \cite{DawidSkene}, iBCC \cite{BCC}, EBCC \cite{li2019exploiting}, LFC \cite{raykar}) and GROUP 2: ZenCrowd (ZC) \cite{zencrowd}, CATD \cite{catd}, GLAD \cite{Whitehill2009WhoseVS}, Majority Vote (MV) )
For each instantiation of the parameters we vary we run the experiments $10$ times, resampling  $N=1000$ data points from MNIST each time.
In the next section we will provide an overview of the results of these experiments.
\section{RESULTS}
\begin{figure}
    \includegraphics[width=0.86\columnwidth]{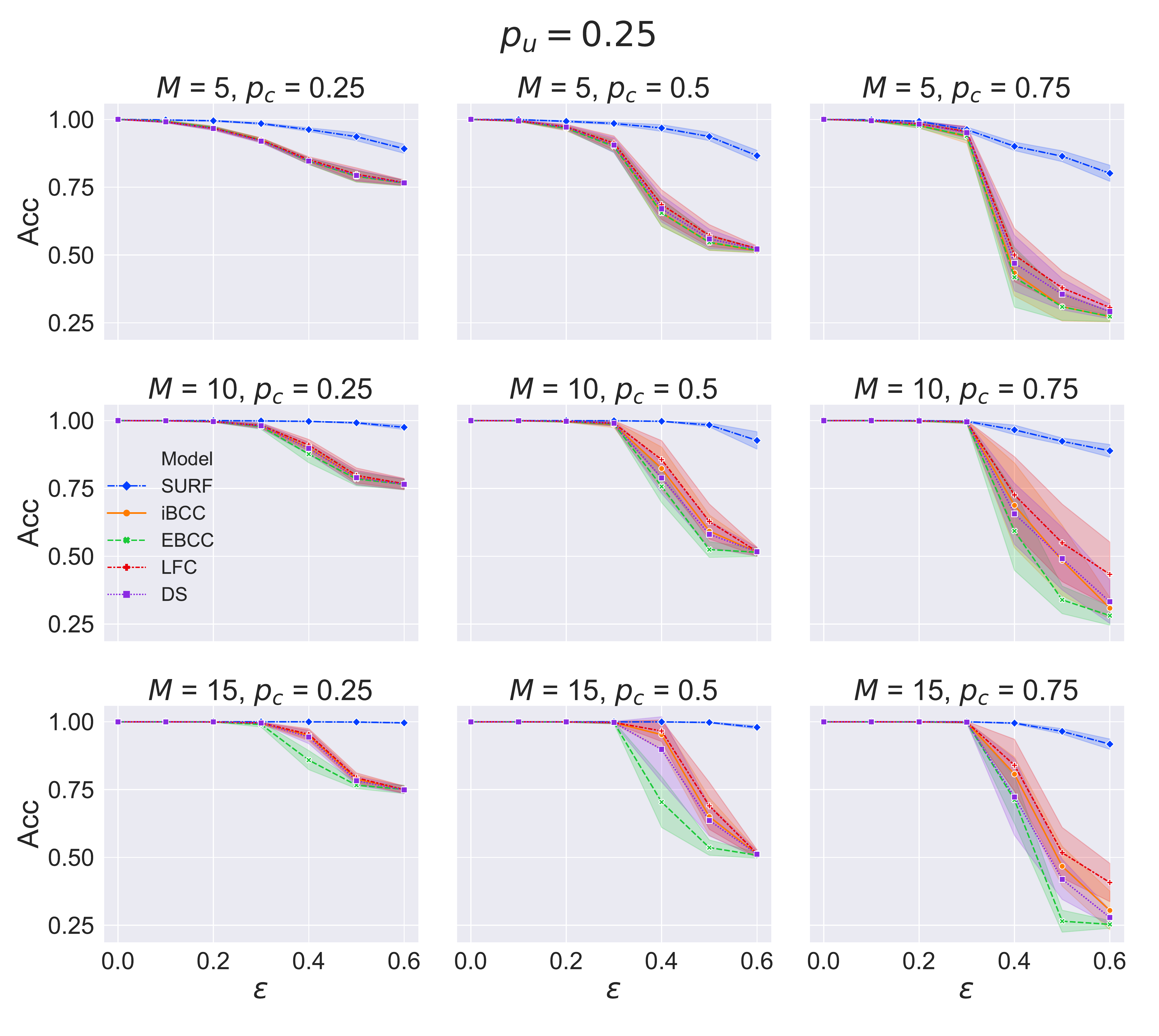}
    \includegraphics[width=0.86\columnwidth]{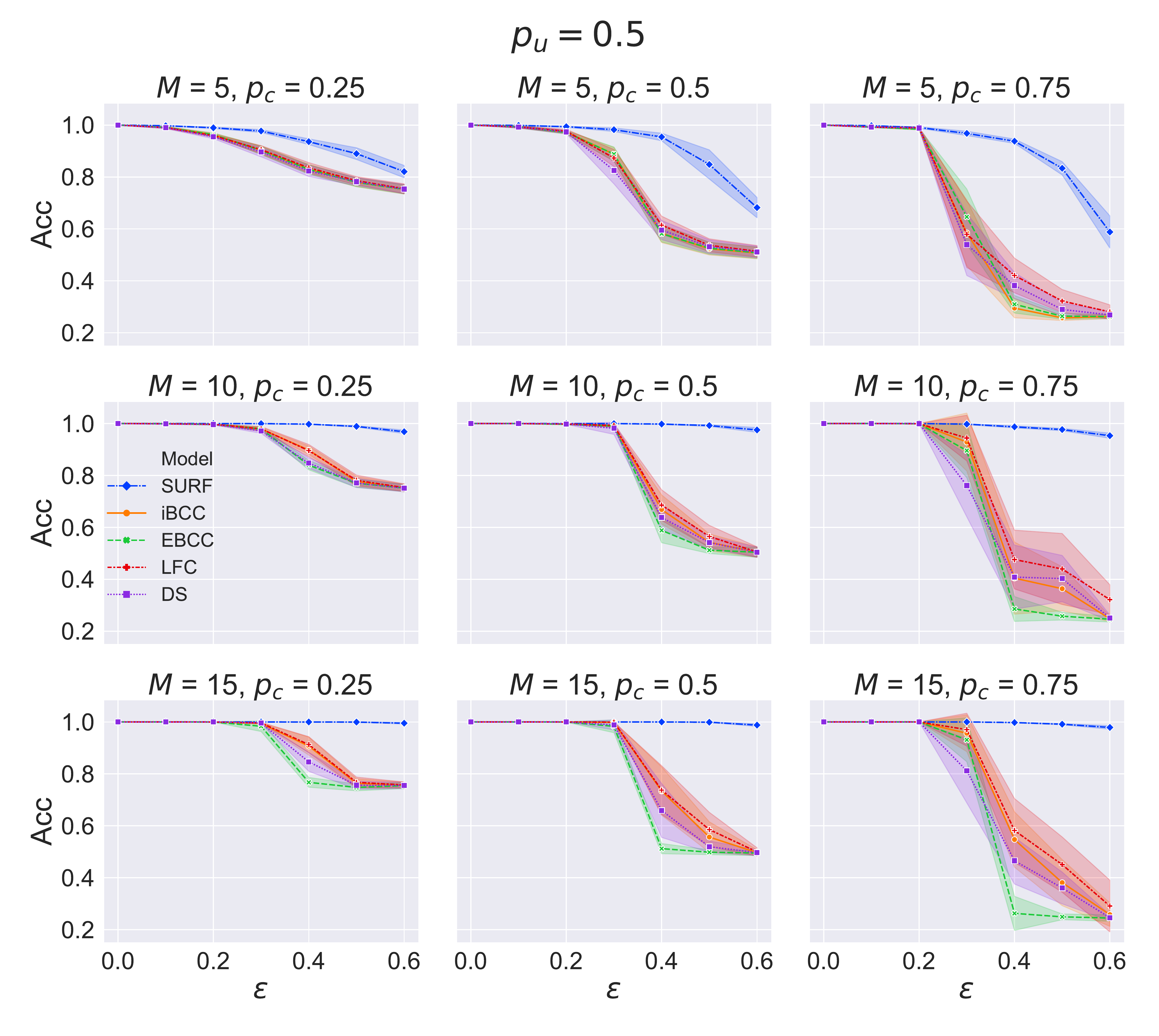}
    \includegraphics[width=0.86\columnwidth]{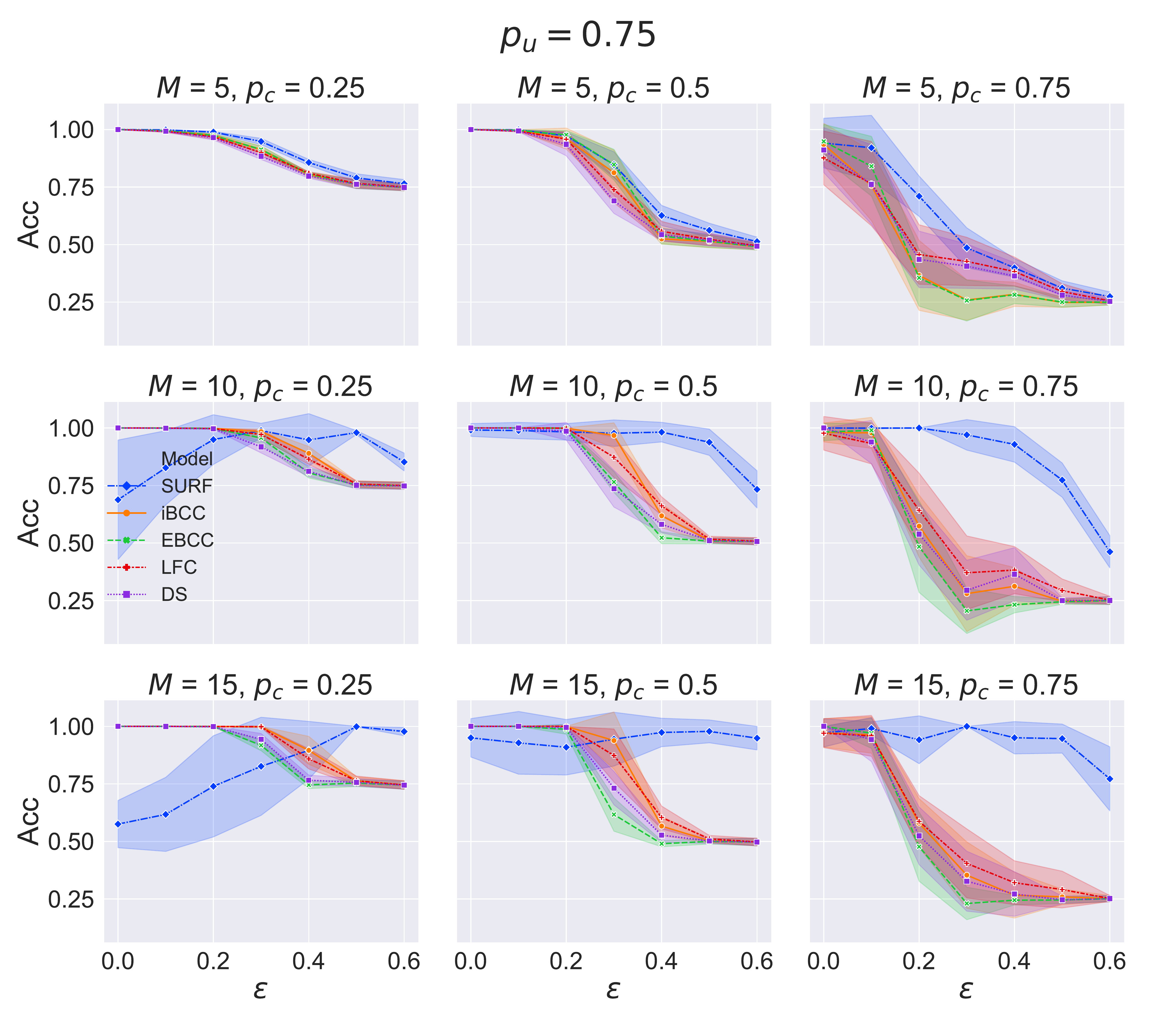}
    \caption{Classification accuracy of GROUP $1$  crowdsourcing algorithms. From top to bottom, we fix user error probability at $p_c=0.25,0.5,0.75$, and we vary number of feedbacking users $M$, classifier error probability $p_c$ and user busyness probability $\epsilon$.}
    \label{fig:figure_one}
\end{figure}

\begin{figure}
    \includegraphics[width=0.86\columnwidth]{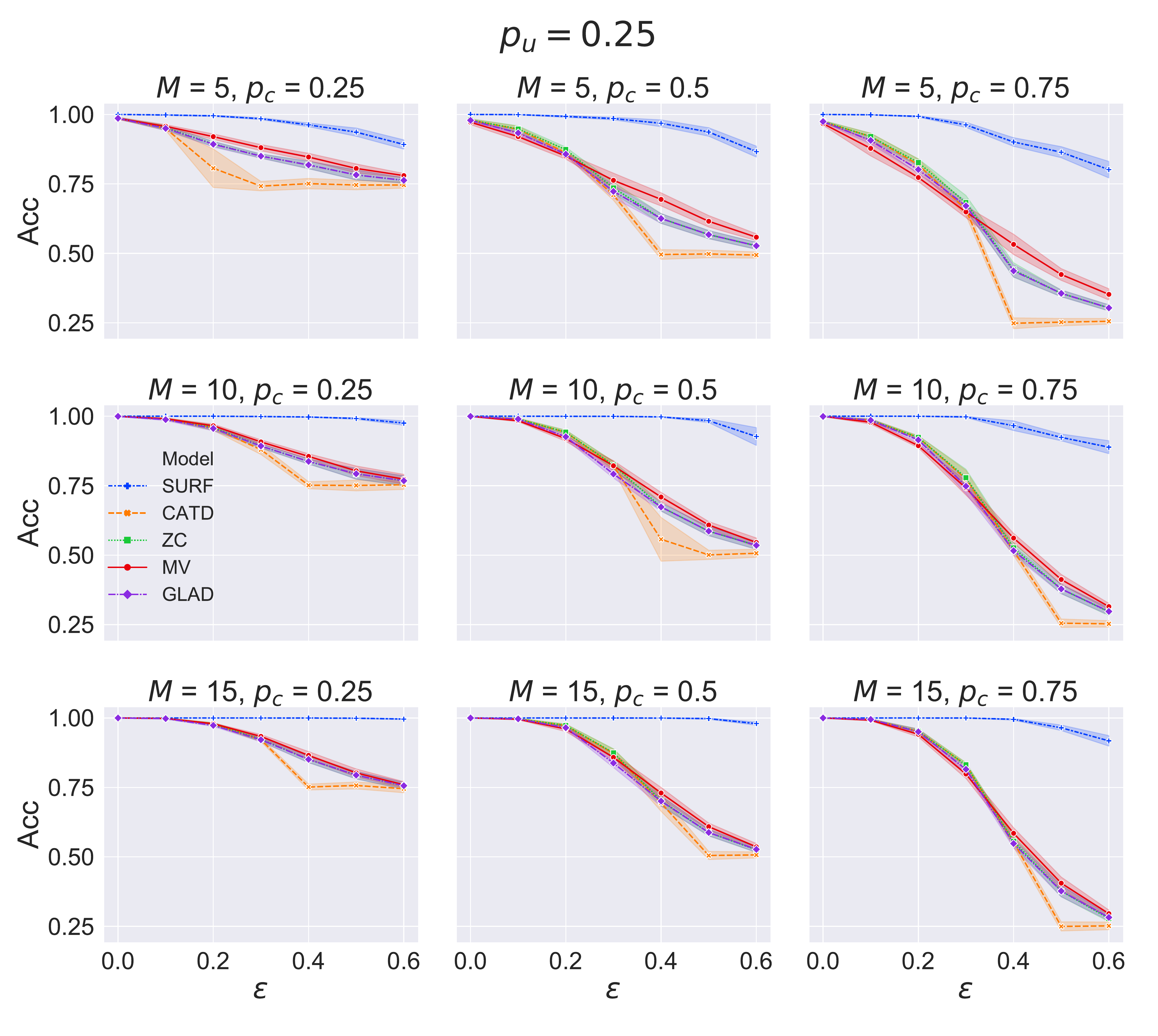}
    \includegraphics[width=0.86\columnwidth]{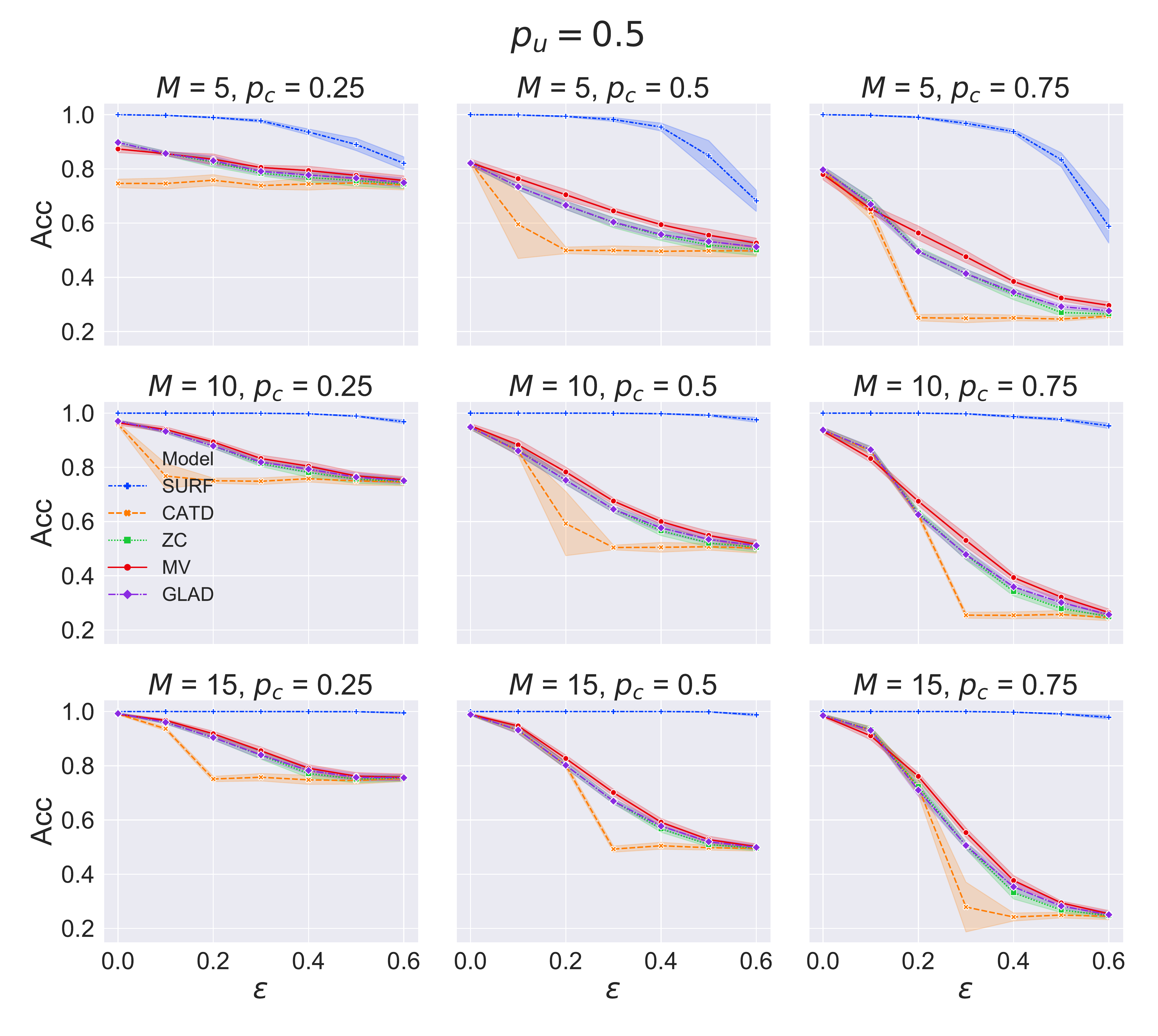}
    \includegraphics[width=0.86\columnwidth]{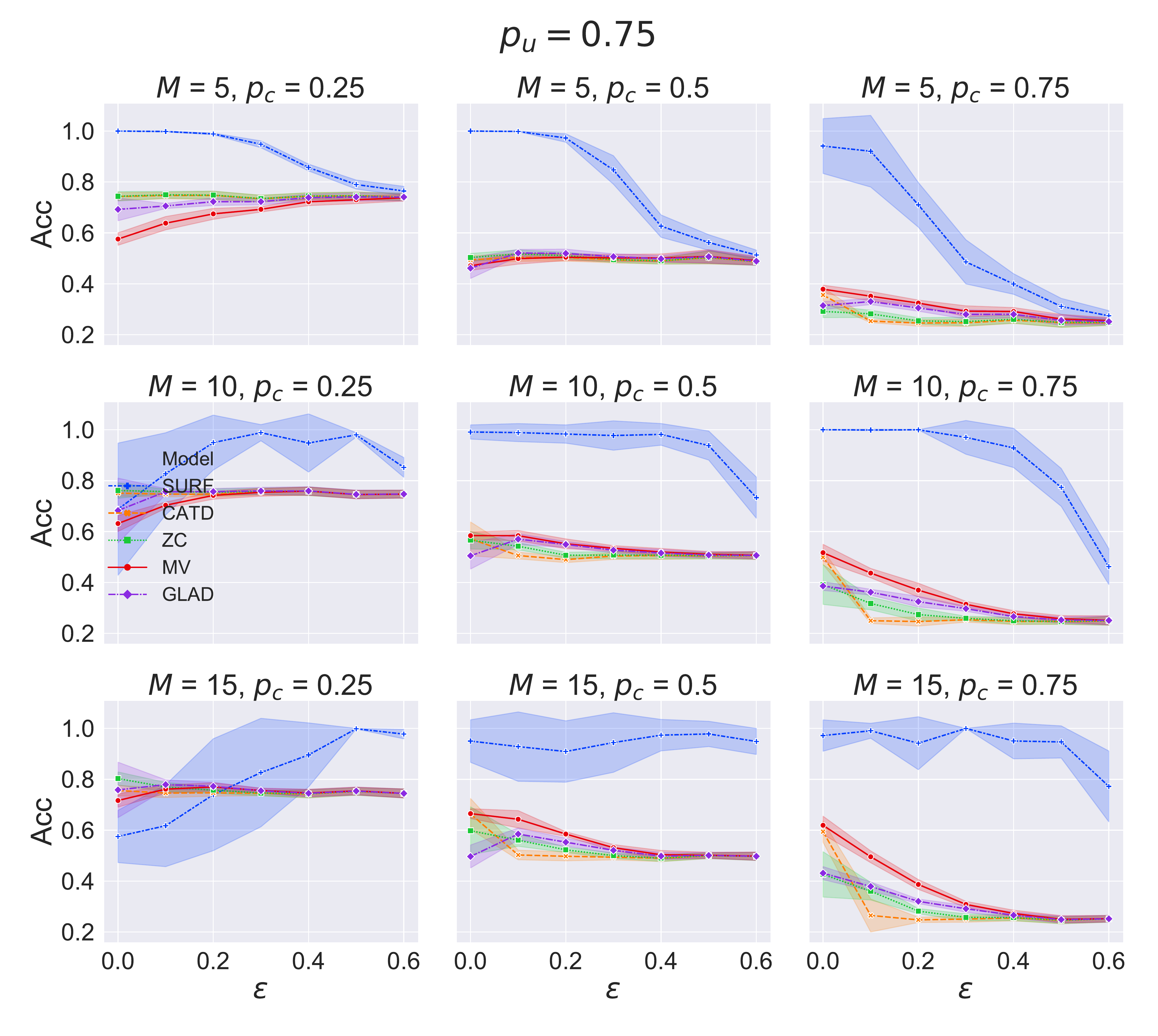}
    \caption{Classification accuracy of GROUP $2$ crowdsourcing algorithms. From top to bottom, we fix user error probability at $p_c=0.25,0.5,0.75$, and we vary number of feedbacking users $M$, classifier error probability $p_c$ and user busyness probability $\epsilon$.}
    \label{fig:figure_two}
\end{figure}

In Figures \ref{fig:figure_one} and \ref{fig:figure_two} we compare the performance of SURF to a number of other crowdsourcing models by plotting the accuracy of each model as a function of user busyness $\epsilon$. For each block of nine plots, we have fixed the user error probability at $p_u$, and varied the classifier error probability, plotting results for $p_c=0.25,0.5,0.75$. We also vary the number of users who have been asked to give feedback on each data point, considering $M=5,10,15$ (recall there are $K=15$ total users).

Figures \ref{fig:figure_one} and \ref{fig:figure_two} validate our intuition on the impact of user busyness on the performance of crowdsourcing algorithms (full results tables are in appendix).
We can extract three insights from the plots:

\emph{(i) SURF outperforms conventional crowdsourcing algorithms when users are busy:}
 In the case where the user busyness probability $\epsilon$ is low, all crowdsourcing algorithms perform well at estimating ground truth. However, as $\epsilon$ increases, we see a degradation in performance in all algorithms but SURF. This degradation in performance is higher when the classifier error probability is higher. This is to be expected, busy users that rarely give feedback on noisy classifiers leads to more noise being injected into the feedback if we assume non-response is agreement.

\emph{(ii) SURF allows us to distinguish between users providing true corrections and users being noisy}:
As discussed previously, a challenge of eliciting feedback from noisy end users is that it is not always clear if users are being noisy in their feedback, or if the classifier itself has an incorrect view on ground truth. A poorly trained classifier that consistently makes mistakes may lead to us filtering out all user feedback as being erroneous, when in fact the users are trying to tell us that the classifier is wrong. However, we see from the figure that when user error probability is low, and classifier error probability is high, SURF's ability to predict ground truth remains high even when busyness probability $\epsilon$ increases. This is vindication of our approach of treating the classifier as simply another user, no `classifier bias' is injected into the process.

\emph{(iii) SURF can operate in sparse feedback environments}:
An additional challenge of crowdsourcing feedback is that not all users in the pool of workers will even see all data points in question. If a team of workers is using the labels provided by the classifier to distribute work to small subgroups of workers (\emph{e.g.} emails labeled as `finance' may simply be forwarded to the accounts department of the company), then we can only rely on receiving feedback from that small subgroup. We simulate such scenarios by varying the number of end users $M$ that \emph{see} each datapoint. The result of this variation is illustrated in the plots, and again, we see that SURF is resilient to variation in this parameter.

We note however, that SURF underperforms in the case where user error probability is high, and classifier error probability is low (see \emph{e.g.} Figure \ref{fig:figure_one} with $p_u=0.75$, $p_c=0.25$). This may be due to our model obtaining a poor estimate of the user busyness probability, which leads to a bad estimate of the ground truth labels (note that this occurs most prominently in cases with low user busyness probability).

In summary, from the results of the experiments we see that all crowdsourcing models perform well when the busyness probability of the team is $\epsilon=0$, but as the busyness probability increases, we see a degradation in performance for all models except our model, SURF. The degradation in performance due to busyness is greater in the cases where classifier error probability is high. This is to be expected, if a user is busy and rarely gives feedback on classifier labels that are mostly incorrect, then estimates based on this feedback are likely to be low quality. Conversely, SURF's performance remains consistently high, and suffers only a mild performance degradation. Finally, SURF is capable of operating with feedback from small subsets of the full user population drawn at random, and doesn't rely on the full team being asked for feedback on every piece of labeled data.

\section{CONCLUSION}

In this work we have considered the problem of eliciting feedback on data points that have been labeled by a supervised learning model. We consider a simulated production scenario: end users are provided with labeled data points, and will provide relabelings of those data points if they feel that the label is incorrect. Furthermore, the end users in our simulation only provide their feedback if they are not \emph{busy}. If a user receives a labeled data point when they are busy, they will not respond and it is assumed that they agree with the label they were sent.

We have shown that the ambiguity surrounding non-response (do they agree with the label, or are they just asleep today?) in the feedback setting can stymie algorithms that are designed for estimating ground truth labels from noisy labels such as Dawid-Skene. To deal with this problem, we presented a refinement of Dawid-Skene that learns what we refer to as each users `busyness probability', and use this to better refine the estimate of the ground truth. The key intuition behind our estimate is to use the uncertainty about our estimate of each data point's label. If a user doesn't give feedback on a data point whose label we are fairly certain about, then the user is probably being non-responsive because they agree with the label. On the other hand, consistent non-response, even for data points about which we are uncertain, will lead us to believe a user has a high busyness probability: they are not engaged in the feedback process.
\begin{acks}
This paper was prepared for informational purposes by the Artificial Intelligence Research group of JPMorgan Chase \& Co and its affiliates (``J.P. Morgan''), and is not a product of the Research Department of J.P. Morgan.  J.P. Morgan makes no representation and warranty whatsoever and disclaims all liability, for the completeness, accuracy or reliability of the information contained herein.  This document is not intended as investment research or investment advice, or a recommendation, offer or solicitation for the purchase or sale of any security, financial instrument, financial product or service, or to be used in any way for evaluating the merits of participating in any transaction, and shall not constitute a solicitation under any jurisdiction or to any person, if such solicitation under such jurisdiction or to such person would be unlawful.   
\end{acks}

\newgeometry{margin=4cm}

\begin{landscape}
\thispagestyle{empty}
\begin{table*}
\scalebox{.47}{
\begin{tabular}{lllllllllllllllllllllllllllll}
\toprule
    & {} & \multicolumn{27}{l}{Acc} \\
    & $M$ & \multicolumn{9}{l}{5} & \multicolumn{9}{l}{10} & \multicolumn{9}{l}{15} \\
    & $p_c$ & \multicolumn{3}{l}{0.25} & \multicolumn{3}{l}{0.50} & \multicolumn{3}{l}{0.75} & \multicolumn{3}{l}{0.25} & \multicolumn{3}{l}{0.50} & \multicolumn{3}{l}{0.75} & \multicolumn{3}{l}{0.25} & \multicolumn{3}{l}{0.50} & \multicolumn{3}{l}{0.75} \\
    & $p_u$ &            0.25 &            0.50 &            0.75 &            0.25 &            0.50 &            0.75 &            0.25 &            0.50 &            0.75 &            0.25 &            0.50 &            0.75 &            0.25 &            0.50 &            0.75 &            0.25 &            0.50 &            0.75 &            0.25 &            0.50 &            0.75 &            0.25 &            0.50 &            0.75 &            0.25 &            0.50 &            0.75 \\
$\epsilon$ & Model &                 &                 &                 &                 &                 &                 &                 &                 &                 &                 &                 &                 &                 &                 &                 &                 &                 &                 &                 &                 &                 &                 &                 &                 &                 &                 &                 \\
\midrule
0.0 & CATD &   0.987$\pm$0.0 &  0.746$\pm$0.02 &  0.743$\pm$0.02 &    0.98$\pm$0.0 &  0.824$\pm$0.01 &  0.494$\pm$0.02 &   0.974$\pm$0.0 &  0.791$\pm$0.02 &  0.356$\pm$0.03 &     1.0$\pm$0.0 &  0.959$\pm$0.01 &  0.751$\pm$0.01 &   0.999$\pm$0.0 &  0.948$\pm$0.01 &  0.571$\pm$0.07 &   0.999$\pm$0.0 &  0.938$\pm$0.01 &  0.499$\pm$0.03 &     1.0$\pm$0.0 &   0.991$\pm$0.0 &  0.756$\pm$0.02 &     1.0$\pm$0.0 &   0.988$\pm$0.0 &  0.664$\pm$0.06 &     1.0$\pm$0.0 &  0.986$\pm$0.01 &  0.594$\pm$0.04 \\
    & DS &     1.0$\pm$0.0 &     1.0$\pm$0.0 &     1.0$\pm$0.0 &     1.0$\pm$0.0 &     1.0$\pm$0.0 &     1.0$\pm$0.0 &     1.0$\pm$0.0 &     1.0$\pm$0.0 &   0.911$\pm$0.1 &     1.0$\pm$0.0 &     1.0$\pm$0.0 &     1.0$\pm$0.0 &     1.0$\pm$0.0 &     1.0$\pm$0.0 &     1.0$\pm$0.0 &     1.0$\pm$0.0 &     1.0$\pm$0.0 &     1.0$\pm$0.0 &     1.0$\pm$0.0 &     1.0$\pm$0.0 &     1.0$\pm$0.0 &     1.0$\pm$0.0 &     1.0$\pm$0.0 &     1.0$\pm$0.0 &     1.0$\pm$0.0 &     1.0$\pm$0.0 &     1.0$\pm$0.0 \\
    & EBCC &     1.0$\pm$0.0 &     1.0$\pm$0.0 &     1.0$\pm$0.0 &     1.0$\pm$0.0 &     1.0$\pm$0.0 &     1.0$\pm$0.0 &     1.0$\pm$0.0 &     1.0$\pm$0.0 &  0.948$\pm$0.07 &     1.0$\pm$0.0 &     1.0$\pm$0.0 &     1.0$\pm$0.0 &     1.0$\pm$0.0 &     1.0$\pm$0.0 &     1.0$\pm$0.0 &     1.0$\pm$0.0 &     1.0$\pm$0.0 &  0.982$\pm$0.04 &     1.0$\pm$0.0 &     1.0$\pm$0.0 &     1.0$\pm$0.0 &     1.0$\pm$0.0 &     1.0$\pm$0.0 &     1.0$\pm$0.0 &     1.0$\pm$0.0 &     1.0$\pm$0.0 &     1.0$\pm$0.0 \\
    & GLAD &   0.986$\pm$0.0 &  0.898$\pm$0.01 &  0.692$\pm$0.04 &   0.979$\pm$0.0 &  0.821$\pm$0.01 &  0.461$\pm$0.04 &   0.974$\pm$0.0 &  0.797$\pm$0.01 &  0.314$\pm$0.02 &     1.0$\pm$0.0 &    0.97$\pm$0.0 &  0.683$\pm$0.13 &     1.0$\pm$0.0 &  0.948$\pm$0.01 &  0.505$\pm$0.05 &   0.999$\pm$0.0 &  0.938$\pm$0.01 &  0.386$\pm$0.02 &     1.0$\pm$0.0 &   0.993$\pm$0.0 &  0.758$\pm$0.11 &     1.0$\pm$0.0 &   0.989$\pm$0.0 &  0.497$\pm$0.05 &     1.0$\pm$0.0 &   0.985$\pm$0.0 &  0.432$\pm$0.03 \\
    & LFC &     1.0$\pm$0.0 &     1.0$\pm$0.0 &     1.0$\pm$0.0 &     1.0$\pm$0.0 &     1.0$\pm$0.0 &     1.0$\pm$0.0 &     1.0$\pm$0.0 &     1.0$\pm$0.0 &  0.877$\pm$0.12 &     1.0$\pm$0.0 &     1.0$\pm$0.0 &     1.0$\pm$0.0 &     1.0$\pm$0.0 &     1.0$\pm$0.0 &     1.0$\pm$0.0 &     1.0$\pm$0.0 &     1.0$\pm$0.0 &  0.977$\pm$0.07 &     1.0$\pm$0.0 &     1.0$\pm$0.0 &     1.0$\pm$0.0 &     1.0$\pm$0.0 &     1.0$\pm$0.0 &     1.0$\pm$0.0 &     1.0$\pm$0.0 &     1.0$\pm$0.0 &   0.97$\pm$0.06 \\
    & MV &   0.986$\pm$0.0 &  0.874$\pm$0.01 &  0.576$\pm$0.02 &  0.972$\pm$0.01 &  0.823$\pm$0.01 &  0.471$\pm$0.02 &  0.967$\pm$0.01 &   0.78$\pm$0.02 &  0.379$\pm$0.02 &   0.999$\pm$0.0 &  0.964$\pm$0.01 &  0.632$\pm$0.03 &   0.999$\pm$0.0 &  0.953$\pm$0.01 &  0.584$\pm$0.01 &     1.0$\pm$0.0 &  0.934$\pm$0.01 &  0.517$\pm$0.03 &     1.0$\pm$0.0 &    0.99$\pm$0.0 &  0.717$\pm$0.03 &     1.0$\pm$0.0 &    0.99$\pm$0.0 &  0.665$\pm$0.02 &     1.0$\pm$0.0 &  0.983$\pm$0.01 &  0.619$\pm$0.04 \\
    & SURF &     1.0$\pm$0.0 &     1.0$\pm$0.0 &     1.0$\pm$0.0 &     1.0$\pm$0.0 &     1.0$\pm$0.0 &     1.0$\pm$0.0 &     1.0$\pm$0.0 &     1.0$\pm$0.0 &  0.941$\pm$0.11 &     1.0$\pm$0.0 &     1.0$\pm$0.0 &  0.688$\pm$0.26 &     1.0$\pm$0.0 &     1.0$\pm$0.0 &  0.991$\pm$0.03 &     1.0$\pm$0.0 &     1.0$\pm$0.0 &     1.0$\pm$0.0 &     1.0$\pm$0.0 &     1.0$\pm$0.0 &   0.576$\pm$0.1 &     1.0$\pm$0.0 &     1.0$\pm$0.0 &   0.95$\pm$0.08 &     1.0$\pm$0.0 &     1.0$\pm$0.0 &  0.972$\pm$0.06 \\
    & ZC &   0.986$\pm$0.0 &  0.898$\pm$0.01 &  0.744$\pm$0.02 &    0.98$\pm$0.0 &  0.822$\pm$0.01 &  0.503$\pm$0.02 &   0.975$\pm$0.0 &  0.796$\pm$0.01 &  0.292$\pm$0.02 &     1.0$\pm$0.0 &    0.97$\pm$0.0 &  0.761$\pm$0.03 &   0.999$\pm$0.0 &  0.948$\pm$0.01 &  0.566$\pm$0.03 &   0.999$\pm$0.0 &  0.938$\pm$0.01 &  0.392$\pm$0.08 &     1.0$\pm$0.0 &   0.993$\pm$0.0 &  0.803$\pm$0.03 &     1.0$\pm$0.0 &   0.988$\pm$0.0 &  0.598$\pm$0.09 &     1.0$\pm$0.0 &   0.986$\pm$0.0 &  0.426$\pm$0.09 \\
    & iBCC &     1.0$\pm$0.0 &     1.0$\pm$0.0 &     1.0$\pm$0.0 &     1.0$\pm$0.0 &     1.0$\pm$0.0 &     1.0$\pm$0.0 &     1.0$\pm$0.0 &     1.0$\pm$0.0 &  0.935$\pm$0.09 &     1.0$\pm$0.0 &     1.0$\pm$0.0 &     1.0$\pm$0.0 &     1.0$\pm$0.0 &     1.0$\pm$0.0 &     1.0$\pm$0.0 &     1.0$\pm$0.0 &     1.0$\pm$0.0 &  0.981$\pm$0.04 &     1.0$\pm$0.0 &     1.0$\pm$0.0 &     1.0$\pm$0.0 &     1.0$\pm$0.0 &     1.0$\pm$0.0 &     1.0$\pm$0.0 &     1.0$\pm$0.0 &     1.0$\pm$0.0 &  0.972$\pm$0.06 \\
0.1 & CATD &   0.95$\pm$0.01 &  0.746$\pm$0.02 &  0.747$\pm$0.01 &  0.947$\pm$0.01 &  0.596$\pm$0.13 &  0.504$\pm$0.02 &   0.92$\pm$0.01 &   0.64$\pm$0.03 &  0.253$\pm$0.01 &    0.99$\pm$0.0 &  0.768$\pm$0.05 &  0.747$\pm$0.01 &    0.99$\pm$0.0 &  0.861$\pm$0.02 &  0.506$\pm$0.01 &   0.986$\pm$0.0 &   0.86$\pm$0.01 &   0.25$\pm$0.01 &   0.998$\pm$0.0 &  0.936$\pm$0.01 &  0.746$\pm$0.02 &   0.997$\pm$0.0 &  0.932$\pm$0.01 &  0.503$\pm$0.02 &   0.995$\pm$0.0 &  0.933$\pm$0.01 &  0.265$\pm$0.06 \\
    & DS &   0.991$\pm$0.0 &    0.99$\pm$0.0 &   0.992$\pm$0.0 &   0.995$\pm$0.0 &   0.993$\pm$0.0 &   0.994$\pm$0.0 &   0.995$\pm$0.0 &   0.992$\pm$0.0 &  0.762$\pm$0.17 &     1.0$\pm$0.0 &   0.999$\pm$0.0 &   0.999$\pm$0.0 &     1.0$\pm$0.0 &     1.0$\pm$0.0 &     1.0$\pm$0.0 &     1.0$\pm$0.0 &     1.0$\pm$0.0 &  0.939$\pm$0.09 &     1.0$\pm$0.0 &     1.0$\pm$0.0 &     1.0$\pm$0.0 &     1.0$\pm$0.0 &     1.0$\pm$0.0 &     1.0$\pm$0.0 &     1.0$\pm$0.0 &     1.0$\pm$0.0 &   0.943$\pm$0.1 \\
    & EBCC &   0.993$\pm$0.0 &   0.991$\pm$0.0 &   0.994$\pm$0.0 &   0.994$\pm$0.0 &   0.993$\pm$0.0 &   0.996$\pm$0.0 &   0.995$\pm$0.0 &   0.992$\pm$0.0 &  0.841$\pm$0.13 &     1.0$\pm$0.0 &     1.0$\pm$0.0 &     1.0$\pm$0.0 &     1.0$\pm$0.0 &     1.0$\pm$0.0 &     1.0$\pm$0.0 &     1.0$\pm$0.0 &     1.0$\pm$0.0 &   0.99$\pm$0.03 &     1.0$\pm$0.0 &     1.0$\pm$0.0 &     1.0$\pm$0.0 &     1.0$\pm$0.0 &     1.0$\pm$0.0 &     1.0$\pm$0.0 &     1.0$\pm$0.0 &     1.0$\pm$0.0 &  0.971$\pm$0.07 \\
    & GLAD &   0.95$\pm$0.01 &  0.857$\pm$0.01 &  0.706$\pm$0.01 &  0.933$\pm$0.01 &  0.734$\pm$0.01 &  0.521$\pm$0.01 &  0.906$\pm$0.01 &   0.67$\pm$0.02 &   0.33$\pm$0.01 &   0.988$\pm$0.0 &  0.933$\pm$0.01 &  0.756$\pm$0.01 &    0.99$\pm$0.0 &  0.861$\pm$0.02 &   0.57$\pm$0.02 &   0.986$\pm$0.0 &  0.865$\pm$0.01 &  0.362$\pm$0.01 &   0.998$\pm$0.0 &   0.96$\pm$0.01 &  0.779$\pm$0.02 &   0.997$\pm$0.0 &  0.932$\pm$0.01 &  0.585$\pm$0.02 &   0.995$\pm$0.0 &  0.931$\pm$0.01 &  0.379$\pm$0.02 \\
    & LFC &   0.991$\pm$0.0 &    0.99$\pm$0.0 &   0.992$\pm$0.0 &   0.994$\pm$0.0 &   0.993$\pm$0.0 &   0.993$\pm$0.0 &   0.996$\pm$0.0 &   0.992$\pm$0.0 &  0.765$\pm$0.18 &     1.0$\pm$0.0 &   0.999$\pm$0.0 &   0.999$\pm$0.0 &     1.0$\pm$0.0 &     1.0$\pm$0.0 &     1.0$\pm$0.0 &     1.0$\pm$0.0 &     1.0$\pm$0.0 &  0.933$\pm$0.09 &     1.0$\pm$0.0 &     1.0$\pm$0.0 &     1.0$\pm$0.0 &     1.0$\pm$0.0 &     1.0$\pm$0.0 &     1.0$\pm$0.0 &     1.0$\pm$0.0 &     1.0$\pm$0.0 &  0.959$\pm$0.09 \\
    & MV &  0.957$\pm$0.01 &  0.856$\pm$0.01 &  0.638$\pm$0.03 &   0.92$\pm$0.01 &  0.764$\pm$0.02 &  0.499$\pm$0.02 &  0.878$\pm$0.03 &  0.652$\pm$0.02 &  0.351$\pm$0.02 &   0.991$\pm$0.0 &   0.94$\pm$0.01 &  0.703$\pm$0.01 &   0.984$\pm$0.0 &  0.883$\pm$0.02 &  0.584$\pm$0.02 &  0.978$\pm$0.01 &  0.832$\pm$0.01 &  0.437$\pm$0.02 &   0.998$\pm$0.0 &   0.968$\pm$0.0 &  0.762$\pm$0.02 &   0.995$\pm$0.0 &  0.947$\pm$0.01 &  0.643$\pm$0.03 &   0.992$\pm$0.0 &   0.91$\pm$0.01 &  0.496$\pm$0.02 \\
    & SURF &   0.998$\pm$0.0 &   0.998$\pm$0.0 &   0.998$\pm$0.0 &   0.999$\pm$0.0 &   0.999$\pm$0.0 &   0.999$\pm$0.0 &   0.999$\pm$0.0 &   0.998$\pm$0.0 &  0.921$\pm$0.14 &     1.0$\pm$0.0 &     1.0$\pm$0.0 &  0.827$\pm$0.16 &     1.0$\pm$0.0 &     1.0$\pm$0.0 &  0.989$\pm$0.04 &     1.0$\pm$0.0 &     1.0$\pm$0.0 &   0.999$\pm$0.0 &     1.0$\pm$0.0 &     1.0$\pm$0.0 &  0.618$\pm$0.16 &     1.0$\pm$0.0 &     1.0$\pm$0.0 &  0.928$\pm$0.14 &     1.0$\pm$0.0 &     1.0$\pm$0.0 &  0.991$\pm$0.03 \\
    & ZC &  0.953$\pm$0.01 &  0.857$\pm$0.01 &   0.75$\pm$0.01 &  0.948$\pm$0.01 &  0.734$\pm$0.01 &  0.517$\pm$0.02 &   0.92$\pm$0.01 &  0.676$\pm$0.02 &  0.282$\pm$0.01 &    0.99$\pm$0.0 &  0.933$\pm$0.01 &  0.757$\pm$0.01 &    0.99$\pm$0.0 &  0.863$\pm$0.02 &  0.542$\pm$0.01 &   0.986$\pm$0.0 &  0.864$\pm$0.01 &  0.317$\pm$0.02 &   0.998$\pm$0.0 &   0.96$\pm$0.01 &   0.77$\pm$0.02 &   0.997$\pm$0.0 &  0.932$\pm$0.01 &  0.561$\pm$0.02 &   0.995$\pm$0.0 &  0.933$\pm$0.01 &  0.361$\pm$0.03 \\
    & iBCC &   0.993$\pm$0.0 &   0.991$\pm$0.0 &   0.994$\pm$0.0 &   0.994$\pm$0.0 &   0.993$\pm$0.0 &   0.997$\pm$0.0 &   0.995$\pm$0.0 &   0.992$\pm$0.0 &  0.756$\pm$0.15 &     1.0$\pm$0.0 &     1.0$\pm$0.0 &     1.0$\pm$0.0 &     1.0$\pm$0.0 &     1.0$\pm$0.0 &     1.0$\pm$0.0 &     1.0$\pm$0.0 &     1.0$\pm$0.0 &  0.979$\pm$0.07 &     1.0$\pm$0.0 &     1.0$\pm$0.0 &     1.0$\pm$0.0 &     1.0$\pm$0.0 &     1.0$\pm$0.0 &     1.0$\pm$0.0 &     1.0$\pm$0.0 &     1.0$\pm$0.0 &  0.963$\pm$0.08 \\
0.2 & CATD &  0.806$\pm$0.07 &  0.758$\pm$0.02 &  0.747$\pm$0.02 &  0.862$\pm$0.01 &    0.5$\pm$0.01 &  0.504$\pm$0.01 &  0.819$\pm$0.02 &  0.251$\pm$0.01 &  0.246$\pm$0.01 &  0.958$\pm$0.01 &  0.751$\pm$0.01 &  0.746$\pm$0.01 &  0.942$\pm$0.01 &  0.593$\pm$0.12 &   0.49$\pm$0.01 &  0.924$\pm$0.01 &  0.626$\pm$0.01 &  0.246$\pm$0.02 &   0.979$\pm$0.0 &  0.751$\pm$0.01 &  0.747$\pm$0.01 &  0.973$\pm$0.01 &  0.802$\pm$0.01 &  0.498$\pm$0.02 &  0.952$\pm$0.01 &  0.721$\pm$0.03 &  0.248$\pm$0.01 \\
    & DS &  0.966$\pm$0.01 &  0.955$\pm$0.01 &  0.965$\pm$0.01 &  0.971$\pm$0.01 &  0.974$\pm$0.01 &  0.936$\pm$0.05 &  0.983$\pm$0.01 &   0.989$\pm$0.0 &  0.436$\pm$0.12 &   0.996$\pm$0.0 &   0.996$\pm$0.0 &   0.996$\pm$0.0 &   0.997$\pm$0.0 &   0.998$\pm$0.0 &  0.986$\pm$0.04 &   0.999$\pm$0.0 &     1.0$\pm$0.0 &  0.539$\pm$0.13 &     1.0$\pm$0.0 &     1.0$\pm$0.0 &   0.999$\pm$0.0 &     1.0$\pm$0.0 &     1.0$\pm$0.0 &  0.996$\pm$0.01 &     1.0$\pm$0.0 &     1.0$\pm$0.0 &  0.525$\pm$0.13 \\
    & EBCC &  0.969$\pm$0.01 &  0.963$\pm$0.01 &  0.974$\pm$0.01 &  0.968$\pm$0.01 &  0.972$\pm$0.01 &  0.977$\pm$0.02 &  0.977$\pm$0.01 &   0.984$\pm$0.0 &  0.356$\pm$0.12 &   0.996$\pm$0.0 &   0.997$\pm$0.0 &   0.998$\pm$0.0 &   0.996$\pm$0.0 &   0.997$\pm$0.0 &   0.999$\pm$0.0 &   0.998$\pm$0.0 &   0.999$\pm$0.0 &   0.484$\pm$0.2 &   0.999$\pm$0.0 &     1.0$\pm$0.0 &     1.0$\pm$0.0 &   0.999$\pm$0.0 &     1.0$\pm$0.0 &  0.987$\pm$0.02 &   0.999$\pm$0.0 &     1.0$\pm$0.0 &  0.477$\pm$0.15 \\
    & GLAD &  0.893$\pm$0.01 &   0.83$\pm$0.02 &  0.723$\pm$0.01 &  0.857$\pm$0.01 &  0.666$\pm$0.02 &  0.519$\pm$0.02 &  0.802$\pm$0.01 &  0.496$\pm$0.01 &  0.305$\pm$0.01 &  0.956$\pm$0.01 &  0.879$\pm$0.01 &  0.757$\pm$0.01 &  0.926$\pm$0.01 &  0.753$\pm$0.02 &  0.549$\pm$0.01 &  0.915$\pm$0.01 &  0.626$\pm$0.01 &  0.325$\pm$0.02 &   0.973$\pm$0.0 &  0.904$\pm$0.01 &  0.773$\pm$0.02 &  0.965$\pm$0.01 &  0.802$\pm$0.01 &  0.553$\pm$0.02 &   0.95$\pm$0.01 &   0.71$\pm$0.02 &  0.321$\pm$0.01 \\
    & LFC &  0.967$\pm$0.01 &  0.959$\pm$0.01 &  0.968$\pm$0.01 &  0.972$\pm$0.01 &  0.976$\pm$0.01 &  0.958$\pm$0.03 &  0.983$\pm$0.01 &    0.99$\pm$0.0 &  0.457$\pm$0.13 &   0.996$\pm$0.0 &   0.997$\pm$0.0 &   0.997$\pm$0.0 &   0.997$\pm$0.0 &   0.998$\pm$0.0 &   0.999$\pm$0.0 &   0.999$\pm$0.0 &     1.0$\pm$0.0 &  0.643$\pm$0.16 &     1.0$\pm$0.0 &     1.0$\pm$0.0 &     1.0$\pm$0.0 &     1.0$\pm$0.0 &     1.0$\pm$0.0 &     1.0$\pm$0.0 &     1.0$\pm$0.0 &     1.0$\pm$0.0 &  0.588$\pm$0.11 \\
    & MV &   0.92$\pm$0.01 &  0.836$\pm$0.02 &  0.675$\pm$0.02 &   0.85$\pm$0.01 &  0.705$\pm$0.02 &  0.504$\pm$0.01 &  0.773$\pm$0.01 &  0.564$\pm$0.03 &  0.324$\pm$0.01 &  0.966$\pm$0.01 &  0.894$\pm$0.01 &  0.742$\pm$0.01 &  0.919$\pm$0.01 &  0.783$\pm$0.02 &  0.552$\pm$0.02 &  0.894$\pm$0.01 &  0.675$\pm$0.02 &   0.37$\pm$0.03 &   0.979$\pm$0.0 &  0.918$\pm$0.01 &   0.77$\pm$0.02 &  0.961$\pm$0.01 &  0.827$\pm$0.01 &  0.585$\pm$0.01 &  0.941$\pm$0.01 &  0.761$\pm$0.01 &  0.387$\pm$0.02 \\
    & SURF &   0.995$\pm$0.0 &    0.99$\pm$0.0 &   0.989$\pm$0.0 &   0.993$\pm$0.0 &   0.994$\pm$0.0 &  0.973$\pm$0.02 &   0.993$\pm$0.0 &   0.991$\pm$0.0 &   0.71$\pm$0.09 &     1.0$\pm$0.0 &     1.0$\pm$0.0 &   0.95$\pm$0.11 &     1.0$\pm$0.0 &     1.0$\pm$0.0 &  0.983$\pm$0.04 &     1.0$\pm$0.0 &     1.0$\pm$0.0 &     1.0$\pm$0.0 &     1.0$\pm$0.0 &     1.0$\pm$0.0 &   0.74$\pm$0.22 &     1.0$\pm$0.0 &     1.0$\pm$0.0 &  0.909$\pm$0.12 &     1.0$\pm$0.0 &     1.0$\pm$0.0 &   0.942$\pm$0.1 \\
    & ZC &  0.894$\pm$0.01 &  0.825$\pm$0.02 &  0.748$\pm$0.02 &  0.874$\pm$0.01 &  0.666$\pm$0.02 &   0.51$\pm$0.01 &  0.827$\pm$0.01 &  0.498$\pm$0.01 &  0.254$\pm$0.01 &  0.959$\pm$0.01 &  0.879$\pm$0.01 &  0.754$\pm$0.01 &  0.943$\pm$0.01 &  0.753$\pm$0.02 &  0.507$\pm$0.01 &  0.924$\pm$0.01 &  0.632$\pm$0.01 &  0.274$\pm$0.03 &   0.979$\pm$0.0 &  0.904$\pm$0.01 &  0.757$\pm$0.01 &  0.973$\pm$0.01 &  0.802$\pm$0.01 &  0.523$\pm$0.01 &  0.952$\pm$0.01 &  0.723$\pm$0.03 &  0.282$\pm$0.01 \\
    & iBCC &  0.969$\pm$0.01 &  0.963$\pm$0.01 &  0.974$\pm$0.01 &  0.968$\pm$0.01 &  0.973$\pm$0.01 &   0.96$\pm$0.05 &  0.978$\pm$0.01 &   0.985$\pm$0.0 &  0.366$\pm$0.15 &   0.996$\pm$0.0 &   0.996$\pm$0.0 &   0.998$\pm$0.0 &   0.996$\pm$0.0 &   0.997$\pm$0.0 &   0.999$\pm$0.0 &   0.998$\pm$0.0 &   0.999$\pm$0.0 &  0.574$\pm$0.13 &   0.999$\pm$0.0 &     1.0$\pm$0.0 &     1.0$\pm$0.0 &   0.999$\pm$0.0 &     1.0$\pm$0.0 &     1.0$\pm$0.0 &   0.999$\pm$0.0 &     1.0$\pm$0.0 &   0.58$\pm$0.11 \\
0.3 & CATD &  0.742$\pm$0.02 &  0.738$\pm$0.01 &  0.734$\pm$0.01 &   0.71$\pm$0.02 &  0.499$\pm$0.02 &  0.494$\pm$0.01 &  0.658$\pm$0.02 &  0.249$\pm$0.02 &  0.247$\pm$0.01 &   0.88$\pm$0.02 &  0.749$\pm$0.01 &  0.756$\pm$0.01 &  0.822$\pm$0.01 &  0.504$\pm$0.01 &  0.504$\pm$0.01 &  0.774$\pm$0.03 &  0.255$\pm$0.01 &  0.254$\pm$0.01 &  0.922$\pm$0.01 &  0.758$\pm$0.01 &  0.745$\pm$0.01 &  0.873$\pm$0.02 &  0.493$\pm$0.01 &  0.494$\pm$0.01 &  0.831$\pm$0.01 &  0.279$\pm$0.09 &  0.251$\pm$0.01 \\
    & DS &   0.92$\pm$0.01 &  0.896$\pm$0.02 &  0.884$\pm$0.01 &  0.905$\pm$0.03 &  0.825$\pm$0.05 &   0.69$\pm$0.05 &  0.952$\pm$0.02 &   0.54$\pm$0.12 &   0.406$\pm$0.1 &  0.981$\pm$0.01 &  0.972$\pm$0.01 &  0.918$\pm$0.03 &   0.99$\pm$0.01 &  0.982$\pm$0.02 &  0.736$\pm$0.08 &   0.996$\pm$0.0 &  0.761$\pm$0.12 &  0.295$\pm$0.13 &   0.996$\pm$0.0 &   0.996$\pm$0.0 &  0.944$\pm$0.03 &   0.998$\pm$0.0 &  0.989$\pm$0.02 &  0.732$\pm$0.07 &   0.999$\pm$0.0 &  0.812$\pm$0.12 &  0.327$\pm$0.13 \\
    & EBCC &  0.924$\pm$0.01 &  0.903$\pm$0.02 &  0.913$\pm$0.01 &    0.9$\pm$0.02 &  0.888$\pm$0.03 &  0.848$\pm$0.07 &  0.939$\pm$0.02 &  0.646$\pm$0.11 &  0.257$\pm$0.09 &  0.978$\pm$0.01 &   0.98$\pm$0.01 &  0.957$\pm$0.03 &  0.987$\pm$0.01 &   0.991$\pm$0.0 &  0.765$\pm$0.05 &  0.994$\pm$0.01 &   0.896$\pm$0.1 &   0.205$\pm$0.1 &   0.99$\pm$0.01 &  0.984$\pm$0.02 &  0.918$\pm$0.02 &   0.997$\pm$0.0 &  0.983$\pm$0.02 &  0.617$\pm$0.07 &   0.998$\pm$0.0 &  0.931$\pm$0.08 &  0.231$\pm$0.07 \\
    & GLAD &   0.85$\pm$0.01 &  0.791$\pm$0.01 &  0.723$\pm$0.01 &  0.723$\pm$0.02 &  0.604$\pm$0.02 &  0.507$\pm$0.01 &  0.671$\pm$0.02 &  0.414$\pm$0.02 &   0.28$\pm$0.02 &  0.893$\pm$0.01 &  0.819$\pm$0.01 &  0.759$\pm$0.01 &  0.792$\pm$0.01 &  0.645$\pm$0.01 &  0.527$\pm$0.01 &  0.748$\pm$0.03 &  0.478$\pm$0.02 &  0.297$\pm$0.01 &  0.922$\pm$0.01 &   0.84$\pm$0.01 &  0.755$\pm$0.01 &  0.838$\pm$0.02 &   0.67$\pm$0.01 &  0.521$\pm$0.01 &  0.815$\pm$0.01 &  0.506$\pm$0.01 &  0.292$\pm$0.01 \\
    & LFC &  0.924$\pm$0.01 &  0.906$\pm$0.02 &    0.9$\pm$0.02 &  0.913$\pm$0.03 &  0.873$\pm$0.03 &  0.739$\pm$0.06 &  0.955$\pm$0.02 &  0.579$\pm$0.13 &   0.427$\pm$0.1 &  0.982$\pm$0.01 &   0.98$\pm$0.01 &  0.972$\pm$0.02 &   0.99$\pm$0.01 &  0.992$\pm$0.01 &  0.873$\pm$0.09 &   0.996$\pm$0.0 &  0.944$\pm$0.09 &  0.371$\pm$0.16 &   0.996$\pm$0.0 &   0.996$\pm$0.0 &   0.999$\pm$0.0 &   0.998$\pm$0.0 &   0.999$\pm$0.0 &  0.874$\pm$0.12 &   0.999$\pm$0.0 &   0.97$\pm$0.06 &  0.405$\pm$0.15 \\
    & MV &   0.88$\pm$0.01 &  0.806$\pm$0.01 &  0.692$\pm$0.01 &  0.763$\pm$0.03 &  0.645$\pm$0.01 &  0.502$\pm$0.01 &  0.649$\pm$0.02 &  0.476$\pm$0.02 &  0.292$\pm$0.02 &  0.908$\pm$0.01 &  0.833$\pm$0.01 &  0.754$\pm$0.02 &  0.822$\pm$0.02 &  0.676$\pm$0.01 &  0.534$\pm$0.01 &  0.742$\pm$0.03 &   0.53$\pm$0.02 &  0.314$\pm$0.01 &  0.934$\pm$0.01 &  0.856$\pm$0.01 &  0.756$\pm$0.01 &  0.858$\pm$0.01 &  0.701$\pm$0.01 &  0.532$\pm$0.01 &  0.798$\pm$0.02 &  0.554$\pm$0.02 &  0.308$\pm$0.01 \\
    & SURF &   0.985$\pm$0.0 &  0.977$\pm$0.01 &  0.948$\pm$0.01 &   0.985$\pm$0.0 &  0.982$\pm$0.01 &  0.847$\pm$0.06 &  0.962$\pm$0.01 &  0.968$\pm$0.01 &  0.486$\pm$0.09 &   0.999$\pm$0.0 &     1.0$\pm$0.0 &  0.988$\pm$0.03 &     1.0$\pm$0.0 &     1.0$\pm$0.0 &  0.977$\pm$0.06 &   0.998$\pm$0.0 &   0.998$\pm$0.0 &   0.97$\pm$0.07 &     1.0$\pm$0.0 &     1.0$\pm$0.0 &  0.826$\pm$0.21 &     1.0$\pm$0.0 &     1.0$\pm$0.0 &  0.944$\pm$0.12 &     1.0$\pm$0.0 &     1.0$\pm$0.0 &     1.0$\pm$0.0 \\
    & ZC &   0.85$\pm$0.01 &  0.784$\pm$0.01 &  0.735$\pm$0.01 &  0.736$\pm$0.02 &  0.602$\pm$0.02 &  0.495$\pm$0.01 &  0.683$\pm$0.03 &  0.413$\pm$0.02 &  0.252$\pm$0.02 &  0.893$\pm$0.01 &  0.814$\pm$0.01 &  0.758$\pm$0.01 &  0.824$\pm$0.01 &  0.645$\pm$0.01 &  0.508$\pm$0.01 &  0.779$\pm$0.03 &  0.477$\pm$0.02 &  0.259$\pm$0.01 &  0.923$\pm$0.01 &  0.838$\pm$0.01 &  0.746$\pm$0.01 &  0.875$\pm$0.01 &   0.67$\pm$0.01 &    0.5$\pm$0.01 &  0.832$\pm$0.01 &  0.505$\pm$0.01 &  0.258$\pm$0.01 \\
    & iBCC &  0.924$\pm$0.01 &  0.903$\pm$0.02 &  0.913$\pm$0.02 &  0.906$\pm$0.02 &  0.888$\pm$0.03 &   0.812$\pm$0.1 &  0.937$\pm$0.02 &  0.585$\pm$0.13 &  0.259$\pm$0.09 &  0.979$\pm$0.01 &  0.979$\pm$0.01 &  0.984$\pm$0.01 &  0.986$\pm$0.01 &   0.991$\pm$0.0 &  0.967$\pm$0.05 &  0.994$\pm$0.01 &  0.929$\pm$0.11 &   0.28$\pm$0.16 &   0.994$\pm$0.0 &   0.994$\pm$0.0 &   0.998$\pm$0.0 &   0.997$\pm$0.0 &   0.999$\pm$0.0 &  0.938$\pm$0.12 &   0.998$\pm$0.0 &  0.956$\pm$0.07 &  0.354$\pm$0.14 \\
0.4 & CATD &  0.751$\pm$0.02 &  0.744$\pm$0.02 &  0.744$\pm$0.01 &  0.496$\pm$0.02 &  0.496$\pm$0.02 &   0.49$\pm$0.01 &  0.248$\pm$0.02 &   0.25$\pm$0.01 &  0.257$\pm$0.01 &  0.752$\pm$0.01 &  0.758$\pm$0.01 &  0.758$\pm$0.02 &  0.557$\pm$0.08 &  0.505$\pm$0.02 &  0.506$\pm$0.01 &  0.516$\pm$0.02 &  0.254$\pm$0.01 &  0.247$\pm$0.01 &  0.752$\pm$0.01 &  0.748$\pm$0.02 &  0.743$\pm$0.02 &  0.694$\pm$0.03 &  0.505$\pm$0.01 &   0.49$\pm$0.01 &  0.553$\pm$0.02 &  0.242$\pm$0.01 &  0.255$\pm$0.01 \\
    & DS &  0.847$\pm$0.01 &  0.823$\pm$0.02 &  0.798$\pm$0.01 &  0.671$\pm$0.05 &  0.596$\pm$0.04 &  0.543$\pm$0.03 &    0.47$\pm$0.1 &  0.382$\pm$0.05 &  0.364$\pm$0.06 &  0.898$\pm$0.02 &  0.848$\pm$0.02 &  0.811$\pm$0.02 &  0.788$\pm$0.06 &  0.638$\pm$0.05 &  0.581$\pm$0.04 &  0.656$\pm$0.11 &  0.408$\pm$0.12 &  0.365$\pm$0.12 &  0.944$\pm$0.02 &  0.846$\pm$0.04 &  0.766$\pm$0.02 &  0.898$\pm$0.12 &   0.659$\pm$0.1 &  0.528$\pm$0.03 &  0.723$\pm$0.14 &  0.466$\pm$0.09 &   0.272$\pm$0.1 \\
    & EBCC &  0.846$\pm$0.01 &  0.829$\pm$0.02 &  0.807$\pm$0.01 &  0.655$\pm$0.05 &  0.582$\pm$0.03 &  0.538$\pm$0.04 &  0.418$\pm$0.11 &   0.31$\pm$0.03 &  0.282$\pm$0.04 &  0.876$\pm$0.03 &   0.84$\pm$0.02 &  0.805$\pm$0.02 &  0.757$\pm$0.06 &  0.589$\pm$0.05 &  0.523$\pm$0.03 &  0.593$\pm$0.15 &  0.286$\pm$0.05 &  0.232$\pm$0.04 &  0.859$\pm$0.04 &  0.767$\pm$0.02 &  0.746$\pm$0.02 &  0.704$\pm$0.09 &  0.512$\pm$0.02 &   0.49$\pm$0.01 &  0.712$\pm$0.09 &  0.263$\pm$0.07 &  0.245$\pm$0.02 \\
    & GLAD &  0.818$\pm$0.01 &  0.777$\pm$0.02 &  0.738$\pm$0.01 &  0.626$\pm$0.02 &  0.558$\pm$0.02 &  0.498$\pm$0.01 &  0.437$\pm$0.02 &  0.346$\pm$0.01 &   0.28$\pm$0.01 &  0.837$\pm$0.01 &  0.793$\pm$0.01 &   0.76$\pm$0.02 &  0.673$\pm$0.01 &  0.577$\pm$0.02 &  0.516$\pm$0.01 &  0.516$\pm$0.02 &  0.359$\pm$0.01 &  0.266$\pm$0.01 &  0.851$\pm$0.01 &  0.783$\pm$0.01 &  0.745$\pm$0.02 &  0.701$\pm$0.02 &  0.577$\pm$0.01 &  0.498$\pm$0.01 &  0.548$\pm$0.02 &  0.353$\pm$0.02 &  0.266$\pm$0.01 \\
    & LFC &  0.852$\pm$0.01 &  0.836$\pm$0.02 &   0.81$\pm$0.01 &  0.686$\pm$0.05 &  0.615$\pm$0.03 &  0.558$\pm$0.04 &     0.5$\pm$0.1 &  0.421$\pm$0.07 &  0.383$\pm$0.06 &   0.91$\pm$0.02 &  0.896$\pm$0.02 &  0.865$\pm$0.02 &  0.856$\pm$0.07 &  0.686$\pm$0.06 &  0.662$\pm$0.04 &  0.726$\pm$0.14 &  0.476$\pm$0.11 &   0.382$\pm$0.1 &  0.954$\pm$0.02 &  0.913$\pm$0.03 &  0.859$\pm$0.04 &  0.966$\pm$0.04 &  0.738$\pm$0.09 &  0.604$\pm$0.05 &    0.84$\pm$0.1 &  0.582$\pm$0.12 &   0.321$\pm$0.1 \\
    & MV &  0.847$\pm$0.01 &  0.794$\pm$0.02 &  0.722$\pm$0.02 &  0.694$\pm$0.02 &  0.594$\pm$0.01 &  0.501$\pm$0.02 &  0.532$\pm$0.04 &  0.385$\pm$0.01 &  0.291$\pm$0.02 &  0.855$\pm$0.01 &  0.804$\pm$0.02 &  0.759$\pm$0.02 &  0.709$\pm$0.02 &    0.6$\pm$0.01 &   0.52$\pm$0.01 &  0.561$\pm$0.02 &  0.394$\pm$0.01 &  0.277$\pm$0.01 &  0.865$\pm$0.01 &  0.791$\pm$0.01 &  0.746$\pm$0.01 &   0.73$\pm$0.02 &  0.592$\pm$0.01 &  0.504$\pm$0.02 &  0.585$\pm$0.02 &  0.377$\pm$0.02 &  0.273$\pm$0.01 \\
    & SURF &  0.962$\pm$0.01 &  0.936$\pm$0.01 &  0.857$\pm$0.01 &  0.968$\pm$0.01 &  0.954$\pm$0.01 &  0.626$\pm$0.04 &  0.901$\pm$0.02 &  0.938$\pm$0.01 &  0.399$\pm$0.04 &   0.997$\pm$0.0 &   0.998$\pm$0.0 &  0.948$\pm$0.11 &   0.998$\pm$0.0 &   0.998$\pm$0.0 &  0.982$\pm$0.04 &  0.966$\pm$0.02 &   0.987$\pm$0.0 &  0.929$\pm$0.08 &     1.0$\pm$0.0 &     1.0$\pm$0.0 &  0.896$\pm$0.13 &     1.0$\pm$0.0 &     1.0$\pm$0.0 &  0.974$\pm$0.06 &   0.995$\pm$0.0 &   0.998$\pm$0.0 &  0.951$\pm$0.07 \\
    & ZC &  0.818$\pm$0.01 &  0.769$\pm$0.02 &  0.745$\pm$0.01 &  0.626$\pm$0.02 &  0.554$\pm$0.02 &   0.49$\pm$0.01 &   0.44$\pm$0.03 &  0.338$\pm$0.02 &  0.261$\pm$0.02 &  0.837$\pm$0.01 &  0.782$\pm$0.02 &  0.758$\pm$0.02 &  0.674$\pm$0.01 &  0.566$\pm$0.02 &  0.507$\pm$0.01 &  0.526$\pm$0.02 &  0.342$\pm$0.02 &   0.25$\pm$0.01 &  0.851$\pm$0.01 &  0.772$\pm$0.02 &  0.743$\pm$0.02 &  0.702$\pm$0.02 &  0.568$\pm$0.01 &   0.49$\pm$0.01 &  0.557$\pm$0.02 &  0.332$\pm$0.02 &  0.257$\pm$0.01 \\
    & iBCC &  0.846$\pm$0.01 &   0.83$\pm$0.02 &  0.804$\pm$0.01 &  0.656$\pm$0.05 &  0.583$\pm$0.04 &  0.525$\pm$0.02 &  0.434$\pm$0.08 &  0.295$\pm$0.04 &  0.284$\pm$0.05 &    0.9$\pm$0.02 &  0.898$\pm$0.02 &   0.89$\pm$0.03 &  0.823$\pm$0.08 &  0.669$\pm$0.06 &  0.618$\pm$0.05 &  0.688$\pm$0.16 &  0.404$\pm$0.14 &  0.312$\pm$0.08 &  0.949$\pm$0.02 &  0.907$\pm$0.03 &  0.896$\pm$0.06 &  0.952$\pm$0.05 &  0.734$\pm$0.09 &  0.567$\pm$0.04 &  0.807$\pm$0.06 &  0.547$\pm$0.11 &   0.267$\pm$0.1 \\
0.5 & CATD &  0.746$\pm$0.02 &  0.748$\pm$0.02 &  0.746$\pm$0.01 &  0.498$\pm$0.01 &  0.498$\pm$0.02 &  0.503$\pm$0.03 &  0.252$\pm$0.01 &  0.246$\pm$0.01 &  0.247$\pm$0.02 &  0.751$\pm$0.02 &   0.75$\pm$0.02 &  0.746$\pm$0.02 &  0.501$\pm$0.02 &  0.507$\pm$0.01 &  0.505$\pm$0.01 &  0.255$\pm$0.02 &  0.257$\pm$0.01 &  0.246$\pm$0.01 &  0.757$\pm$0.01 &  0.745$\pm$0.01 &  0.752$\pm$0.02 &  0.505$\pm$0.01 &  0.498$\pm$0.01 &    0.5$\pm$0.01 &   0.25$\pm$0.02 &  0.249$\pm$0.01 &  0.246$\pm$0.01 \\
    & DS &  0.794$\pm$0.02 &  0.782$\pm$0.02 &  0.762$\pm$0.02 &   0.56$\pm$0.04 &  0.532$\pm$0.03 &   0.52$\pm$0.03 &  0.355$\pm$0.06 &   0.29$\pm$0.02 &   0.28$\pm$0.02 &   0.79$\pm$0.03 &  0.772$\pm$0.02 &  0.751$\pm$0.02 &  0.581$\pm$0.05 &  0.541$\pm$0.02 &  0.511$\pm$0.01 &  0.491$\pm$0.12 &  0.403$\pm$0.09 &   0.25$\pm$0.01 &  0.783$\pm$0.02 &  0.756$\pm$0.02 &  0.757$\pm$0.02 &  0.636$\pm$0.06 &  0.519$\pm$0.02 &  0.502$\pm$0.01 &  0.419$\pm$0.07 &  0.361$\pm$0.06 &  0.246$\pm$0.02 \\
    & EBCC &   0.79$\pm$0.02 &   0.78$\pm$0.02 &  0.762$\pm$0.02 &  0.547$\pm$0.03 &  0.526$\pm$0.02 &  0.515$\pm$0.03 &  0.309$\pm$0.05 &  0.264$\pm$0.01 &   0.25$\pm$0.02 &  0.785$\pm$0.03 &   0.77$\pm$0.02 &  0.752$\pm$0.02 &  0.525$\pm$0.03 &  0.512$\pm$0.01 &  0.508$\pm$0.01 &  0.339$\pm$0.05 &  0.258$\pm$0.01 &  0.245$\pm$0.01 &  0.766$\pm$0.01 &  0.748$\pm$0.01 &  0.754$\pm$0.02 &  0.536$\pm$0.03 &  0.498$\pm$0.01 &    0.5$\pm$0.01 &  0.265$\pm$0.04 &  0.249$\pm$0.01 &  0.246$\pm$0.01 \\
    & GLAD &  0.782$\pm$0.02 &  0.766$\pm$0.02 &  0.741$\pm$0.02 &  0.568$\pm$0.01 &  0.532$\pm$0.02 &  0.506$\pm$0.03 &  0.356$\pm$0.01 &  0.292$\pm$0.01 &  0.256$\pm$0.02 &  0.793$\pm$0.02 &  0.764$\pm$0.01 &  0.746$\pm$0.02 &  0.586$\pm$0.02 &  0.534$\pm$0.02 &  0.508$\pm$0.01 &  0.378$\pm$0.02 &  0.301$\pm$0.02 &  0.253$\pm$0.01 &  0.794$\pm$0.01 &  0.757$\pm$0.01 &  0.754$\pm$0.02 &  0.588$\pm$0.01 &   0.52$\pm$0.01 &  0.501$\pm$0.01 &  0.377$\pm$0.02 &  0.282$\pm$0.01 &  0.249$\pm$0.01 \\
    & LFC &  0.798$\pm$0.02 &  0.786$\pm$0.02 &  0.766$\pm$0.02 &  0.572$\pm$0.04 &  0.537$\pm$0.02 &  0.523$\pm$0.03 &  0.379$\pm$0.06 &  0.322$\pm$0.04 &  0.296$\pm$0.03 &  0.798$\pm$0.03 &  0.781$\pm$0.02 &  0.756$\pm$0.01 &  0.628$\pm$0.06 &  0.565$\pm$0.04 &  0.517$\pm$0.01 &  0.549$\pm$0.14 &   0.44$\pm$0.14 &  0.294$\pm$0.05 &  0.794$\pm$0.02 &  0.768$\pm$0.02 &  0.764$\pm$0.02 &   0.69$\pm$0.09 &  0.586$\pm$0.07 &  0.511$\pm$0.01 &  0.517$\pm$0.09 &  0.451$\pm$0.11 &  0.291$\pm$0.08 \\
    & MV &  0.806$\pm$0.02 &  0.776$\pm$0.02 &   0.73$\pm$0.02 &  0.616$\pm$0.02 &  0.556$\pm$0.02 &  0.507$\pm$0.03 &  0.424$\pm$0.02 &  0.324$\pm$0.01 &  0.261$\pm$0.02 &  0.803$\pm$0.02 &  0.768$\pm$0.01 &  0.745$\pm$0.02 &  0.608$\pm$0.01 &  0.549$\pm$0.02 &  0.511$\pm$0.01 &  0.412$\pm$0.02 &  0.321$\pm$0.02 &  0.258$\pm$0.01 &  0.802$\pm$0.01 &  0.762$\pm$0.01 &  0.755$\pm$0.01 &  0.609$\pm$0.01 &  0.528$\pm$0.01 &  0.502$\pm$0.01 &  0.406$\pm$0.02 &  0.294$\pm$0.01 &   0.25$\pm$0.01 \\
    & SURF &  0.936$\pm$0.01 &   0.89$\pm$0.02 &   0.79$\pm$0.02 &  0.937$\pm$0.02 &  0.848$\pm$0.06 &  0.562$\pm$0.03 &  0.864$\pm$0.02 &  0.834$\pm$0.03 &  0.311$\pm$0.03 &   0.992$\pm$0.0 &   0.989$\pm$0.0 &   0.98$\pm$0.01 &  0.984$\pm$0.01 &   0.992$\pm$0.0 &  0.938$\pm$0.06 &  0.924$\pm$0.01 &   0.977$\pm$0.0 &  0.774$\pm$0.07 &   0.999$\pm$0.0 &     1.0$\pm$0.0 &   0.999$\pm$0.0 &   0.998$\pm$0.0 &   0.999$\pm$0.0 &  0.978$\pm$0.05 &  0.965$\pm$0.01 &   0.991$\pm$0.0 &  0.947$\pm$0.06 \\
    & ZC &  0.782$\pm$0.02 &  0.756$\pm$0.02 &  0.746$\pm$0.01 &  0.568$\pm$0.01 &  0.519$\pm$0.02 &  0.504$\pm$0.03 &  0.356$\pm$0.01 &   0.27$\pm$0.01 &  0.248$\pm$0.02 &  0.793$\pm$0.02 &  0.757$\pm$0.02 &  0.746$\pm$0.02 &  0.586$\pm$0.02 &  0.521$\pm$0.02 &  0.505$\pm$0.01 &  0.378$\pm$0.02 &  0.279$\pm$0.01 &  0.247$\pm$0.01 &  0.794$\pm$0.01 &  0.751$\pm$0.01 &  0.753$\pm$0.01 &  0.588$\pm$0.01 &  0.509$\pm$0.01 &    0.5$\pm$0.01 &  0.377$\pm$0.02 &  0.268$\pm$0.01 &  0.246$\pm$0.01 \\
    & iBCC &  0.791$\pm$0.02 &   0.78$\pm$0.02 &  0.762$\pm$0.02 &  0.546$\pm$0.03 &  0.524$\pm$0.02 &  0.514$\pm$0.03 &   0.31$\pm$0.05 &  0.257$\pm$0.01 &  0.249$\pm$0.02 &   0.79$\pm$0.03 &  0.777$\pm$0.02 &  0.755$\pm$0.01 &  0.593$\pm$0.06 &  0.542$\pm$0.03 &  0.512$\pm$0.01 &  0.484$\pm$0.13 &  0.364$\pm$0.08 &  0.247$\pm$0.01 &  0.788$\pm$0.02 &  0.764$\pm$0.02 &  0.763$\pm$0.02 &   0.65$\pm$0.07 &  0.557$\pm$0.06 &  0.504$\pm$0.01 &  0.467$\pm$0.07 &  0.381$\pm$0.09 &  0.259$\pm$0.03 \\
0.6 & CATD &  0.746$\pm$0.01 &  0.741$\pm$0.02 &  0.741$\pm$0.01 &  0.494$\pm$0.01 &  0.498$\pm$0.02 &  0.486$\pm$0.01 &  0.256$\pm$0.01 &  0.256$\pm$0.01 &  0.247$\pm$0.01 &  0.753$\pm$0.02 &  0.746$\pm$0.01 &  0.747$\pm$0.02 &  0.507$\pm$0.01 &  0.502$\pm$0.02 &  0.505$\pm$0.02 &  0.253$\pm$0.01 &  0.246$\pm$0.01 &  0.249$\pm$0.02 &  0.746$\pm$0.01 &  0.754$\pm$0.01 &  0.744$\pm$0.02 &  0.507$\pm$0.01 &  0.495$\pm$0.01 &  0.497$\pm$0.02 &  0.252$\pm$0.01 &  0.245$\pm$0.01 &  0.252$\pm$0.01 \\
    & DS &  0.766$\pm$0.01 &  0.754$\pm$0.02 &  0.748$\pm$0.01 &  0.522$\pm$0.01 &  0.511$\pm$0.02 &  0.493$\pm$0.01 &  0.292$\pm$0.03 &  0.269$\pm$0.01 &  0.253$\pm$0.01 &  0.765$\pm$0.02 &  0.751$\pm$0.01 &  0.748$\pm$0.01 &  0.516$\pm$0.02 &  0.505$\pm$0.02 &  0.507$\pm$0.01 &  0.332$\pm$0.08 &  0.251$\pm$0.01 &  0.251$\pm$0.02 &  0.749$\pm$0.01 &  0.756$\pm$0.01 &  0.745$\pm$0.02 &  0.512$\pm$0.01 &  0.497$\pm$0.01 &  0.498$\pm$0.02 &  0.279$\pm$0.03 &  0.246$\pm$0.01 &  0.252$\pm$0.01 \\
    & EBCC &  0.765$\pm$0.01 &  0.752$\pm$0.02 &  0.748$\pm$0.01 &  0.517$\pm$0.01 &  0.508$\pm$0.02 &  0.491$\pm$0.01 &  0.274$\pm$0.02 &  0.263$\pm$0.01 &   0.25$\pm$0.01 &  0.765$\pm$0.02 &  0.752$\pm$0.01 &  0.749$\pm$0.02 &  0.515$\pm$0.02 &  0.504$\pm$0.02 &  0.507$\pm$0.01 &  0.281$\pm$0.04 &  0.246$\pm$0.01 &   0.25$\pm$0.02 &  0.751$\pm$0.02 &  0.755$\pm$0.01 &  0.745$\pm$0.02 &  0.508$\pm$0.01 &  0.495$\pm$0.01 &  0.497$\pm$0.02 &  0.253$\pm$0.01 &  0.245$\pm$0.01 &  0.252$\pm$0.01 \\
    & GLAD &  0.763$\pm$0.01 &  0.749$\pm$0.02 &  0.741$\pm$0.01 &  0.527$\pm$0.01 &  0.513$\pm$0.02 &  0.489$\pm$0.02 &  0.304$\pm$0.01 &  0.276$\pm$0.01 &  0.251$\pm$0.01 &  0.768$\pm$0.02 &   0.75$\pm$0.01 &  0.747$\pm$0.02 &  0.535$\pm$0.01 &  0.511$\pm$0.02 &  0.506$\pm$0.01 &  0.297$\pm$0.01 &  0.257$\pm$0.01 &  0.251$\pm$0.02 &  0.756$\pm$0.01 &  0.756$\pm$0.01 &  0.745$\pm$0.02 &  0.527$\pm$0.01 &  0.499$\pm$0.01 &  0.498$\pm$0.02 &  0.282$\pm$0.01 &   0.25$\pm$0.01 &  0.252$\pm$0.01 \\
    & LFC &  0.767$\pm$0.01 &  0.755$\pm$0.02 &   0.75$\pm$0.02 &  0.524$\pm$0.01 &  0.514$\pm$0.02 &  0.495$\pm$0.01 &  0.306$\pm$0.03 &   0.28$\pm$0.03 &  0.256$\pm$0.01 &  0.767$\pm$0.02 &  0.753$\pm$0.01 &   0.75$\pm$0.02 &  0.518$\pm$0.02 &  0.505$\pm$0.02 &  0.508$\pm$0.02 &  0.433$\pm$0.12 &  0.322$\pm$0.06 &  0.252$\pm$0.02 &   0.75$\pm$0.01 &  0.757$\pm$0.01 &  0.746$\pm$0.02 &  0.514$\pm$0.01 &    0.5$\pm$0.02 &  0.498$\pm$0.02 &  0.408$\pm$0.07 &   0.291$\pm$0.1 &  0.252$\pm$0.01 \\
    & MV &   0.78$\pm$0.01 &  0.757$\pm$0.02 &  0.738$\pm$0.01 &  0.558$\pm$0.01 &  0.526$\pm$0.02 &  0.491$\pm$0.02 &  0.352$\pm$0.02 &  0.296$\pm$0.01 &  0.255$\pm$0.01 &  0.773$\pm$0.02 &  0.754$\pm$0.01 &  0.747$\pm$0.02 &  0.546$\pm$0.02 &  0.516$\pm$0.02 &  0.507$\pm$0.01 &  0.315$\pm$0.01 &  0.264$\pm$0.01 &  0.252$\pm$0.02 &   0.76$\pm$0.01 &  0.758$\pm$0.01 &  0.745$\pm$0.02 &  0.536$\pm$0.01 &  0.502$\pm$0.01 &  0.498$\pm$0.02 &  0.296$\pm$0.01 &  0.254$\pm$0.01 &  0.252$\pm$0.01 \\
    & SURF &  0.892$\pm$0.02 &  0.821$\pm$0.02 &  0.765$\pm$0.02 &  0.866$\pm$0.02 &  0.682$\pm$0.04 &  0.513$\pm$0.02 &  0.801$\pm$0.03 &  0.588$\pm$0.06 &  0.274$\pm$0.02 &  0.975$\pm$0.01 &  0.968$\pm$0.01 &  0.852$\pm$0.04 &  0.927$\pm$0.03 &  0.975$\pm$0.01 &  0.733$\pm$0.08 &  0.889$\pm$0.02 &  0.953$\pm$0.01 &  0.462$\pm$0.07 &   0.996$\pm$0.0 &   0.995$\pm$0.0 &  0.978$\pm$0.02 &   0.98$\pm$0.01 &  0.988$\pm$0.01 &  0.949$\pm$0.05 &  0.918$\pm$0.02 &  0.979$\pm$0.01 &  0.772$\pm$0.14 \\
    & ZC &  0.763$\pm$0.01 &  0.744$\pm$0.02 &  0.741$\pm$0.01 &  0.527$\pm$0.01 &  0.502$\pm$0.02 &  0.487$\pm$0.01 &  0.304$\pm$0.01 &  0.265$\pm$0.01 &  0.248$\pm$0.01 &  0.768$\pm$0.02 &  0.748$\pm$0.01 &  0.747$\pm$0.02 &  0.535$\pm$0.01 &  0.506$\pm$0.02 &  0.506$\pm$0.01 &  0.296$\pm$0.01 &   0.25$\pm$0.01 &  0.249$\pm$0.02 &  0.756$\pm$0.01 &  0.754$\pm$0.01 &  0.744$\pm$0.02 &  0.526$\pm$0.01 &  0.496$\pm$0.01 &  0.497$\pm$0.02 &  0.278$\pm$0.01 &  0.247$\pm$0.01 &  0.252$\pm$0.01 \\
    & iBCC &  0.765$\pm$0.01 &  0.752$\pm$0.02 &  0.748$\pm$0.01 &  0.516$\pm$0.01 &  0.507$\pm$0.02 &  0.491$\pm$0.01 &  0.272$\pm$0.02 &  0.262$\pm$0.01 &  0.249$\pm$0.01 &  0.764$\pm$0.02 &  0.752$\pm$0.01 &   0.75$\pm$0.02 &  0.516$\pm$0.02 &  0.505$\pm$0.02 &  0.507$\pm$0.01 &  0.308$\pm$0.04 &  0.248$\pm$0.01 &   0.25$\pm$0.02 &   0.75$\pm$0.01 &  0.757$\pm$0.01 &  0.746$\pm$0.02 &  0.513$\pm$0.01 &  0.497$\pm$0.01 &  0.498$\pm$0.02 &  0.304$\pm$0.07 &  0.258$\pm$0.05 &  0.252$\pm$0.01 \\
\bottomrule
\end{tabular}
}
\end{table*}
\end{landscape}
\restoregeometry
\end{document}